\documentclass{article}
\usepackage{thestyle,times}

\usepackage[utf8]{inputenc}
\usepackage[T1]{fontenc}
\usepackage{hyperref}
\usepackage{url}
\usepackage{booktabs}
\usepackage{amsfonts}
\usepackage{amsmath}
\usepackage{nicefrac}
\usepackage{microtype}
\usepackage{xcolor}
\usepackage{graphicx}
\usepackage{float}
\usepackage{placeins}
\usepackage{subcaption}
\usepackage{multirow}
\usepackage{tabularx}
\usepackage{wrapfig}
\usepackage{makecell}

\usepackage{lipsum}

\title{Structural Affinity: An Algorithm for \\ Coherent Cage-Based Shape Design}

\author{
Aviraj Newatia \And
Shengan Yang \And
Hantao Zhong \And
Rika Antonova\\[0.5em]
University of Cambridge\\
}

\preprintcopy

\begin{document}
\maketitle

\begin{abstract}
Cage-based deformation (CBD) parametrises 3D geometry through a small set of control handles, effectively enabling a variety of tasks from animation and interactive modelling to tool-based co-design, neural deformation transfer, and aerodynamic shape optimization.
Many optimization approaches relying on CBD employ regularisation strategies to reduce the distortion of the deformed mesh by encouraging smooth, regular cage deformations.
These techniques offer soft guidance to an optimization algorithm towards desirable deformations, but compete with the task objective rather than changing the deformation parameterization itself.
In this work we introduce \textit{Structural Affinity} for CBD, an explicit regularisation method which enables smooth deformations through control cages by exploiting a two-cage structure to couple cage-handle motion directly.
\textit{Structural Affinity} precomputes an affinity matrix using mean-value coordinates, binding the inner control-cage to a coarser outer-cage. This matrix defines how much a cage handle displacement propagates to neighbouring handles which substantially decreases and often eliminates cage self-intersections, reduces mesh distortion, and adds linear overhead at deformation time.
Across robotic control--morphology co-design, source-to-target neural deformation transfer, and aerodynamic wing shape optimization, \textit{Structural Affinity} produces more regular deformations and substantially reduces reliance on hand-designed structural loss terms. We report consistent gains on three complementary diagnostics: the cotangent Laplacian, a proxy for surface smoothness; cage distortion, capturing whether control handles move coherently; and surface roughness, which measures undesirable oscillation in wing geometry.
\let\thefootnote\relax\footnotetext{\texttt{\{aviraj.newatia,shengan.yang,hantao.zhong,rika.antonova\}@cst.cam.ac.uk}}

\end{abstract}

%========================================================
\section{Introduction}

Algorithmic design and deformation of 3D shapes is employed in computer vision, graphics, and robotics. 
However, searching the space of arbitrary shapes causes a combinatorial explosion with many degrees of freedom to optimize, and vast regions with irregular, distorted, or degenerate shapes.

Reduced-parameterization modeling techniques such as Cage-Based Deformation~\citep{gonzalez_hidalgo_cage_2013} (CBD) and Free-Form Deformation~\citep{sederberg_free-form_1986} (FFD) are typically used to reduce the search space over shapes.
By preserving desirable attributes of the underlying shape (e.g. mesh topology), such methods reduce the degrees of freedom for shape deformation. This implicitly reducing the complexity of the search space, as compared to directly predicting deltas on mesh vertices, or predicting the vertices and edges themselves. 
Operating search algorithms or optimization under such parametrizations restricts the search space to a region where desirable attributes are preserved (with respect to the base mesh). 
Furthermore, in the case of CBD, different classes of attributes can be preserved depending on the weight computation method employed. For instance, Somigliana coordinates~\citep{chen_somigliana_2023} produce elastic-style deformations.

\begin{wrapfigure}{r}{0.6\textwidth}
\vspace{-11px}
\includegraphics[width=0.6\textwidth]{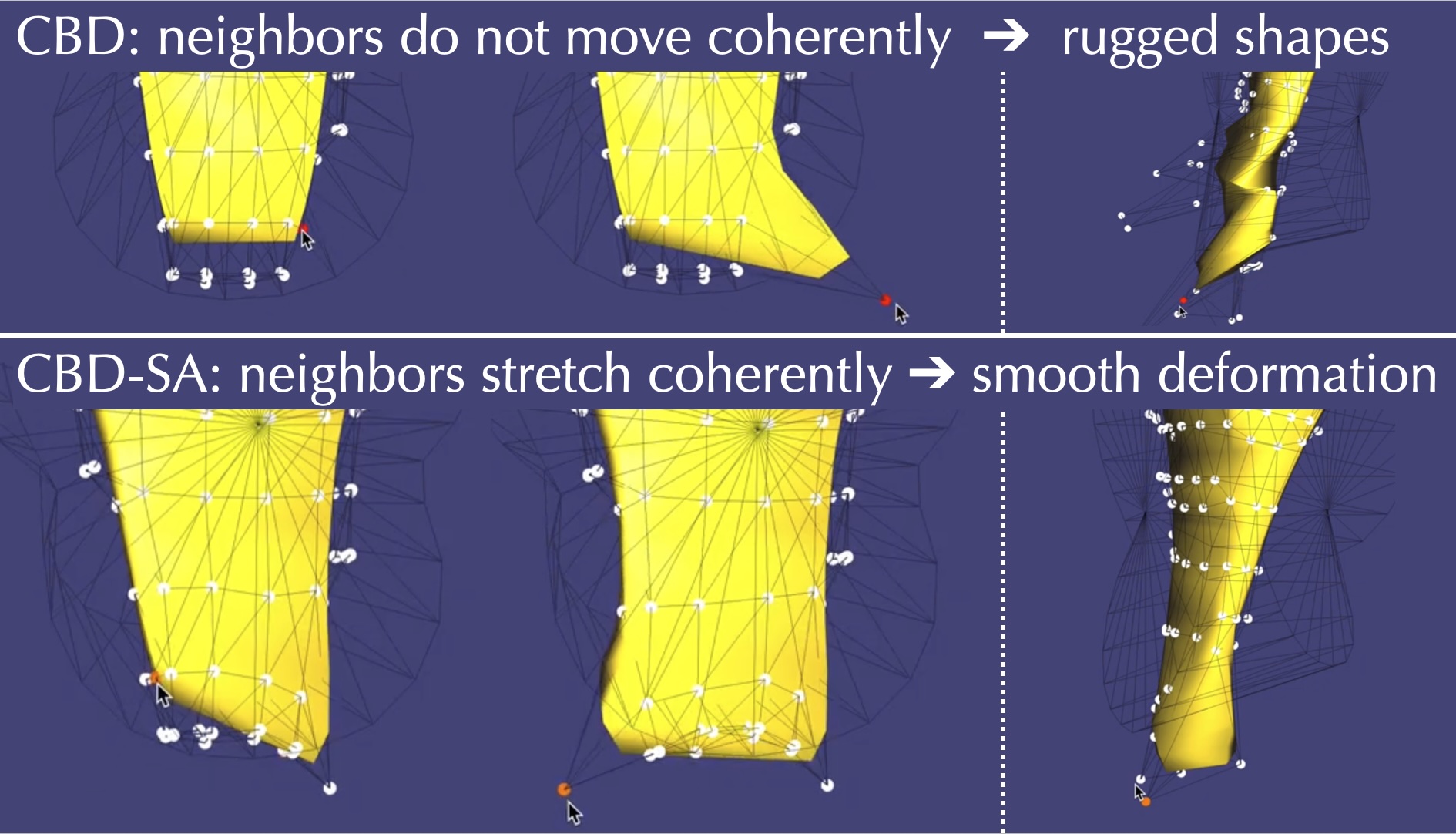}
\vspace{-15px}
\caption{Top: CBD; bottom: CBD-SA.}
\label{fig:overview}
\vspace{-20px}
\end{wrapfigure}

CBD, however, is still highly sensitive to the movement of its control handles. Poor control of cage handles can lead to distorted, "spiky", and degenerate shapes (e.g. Figure~\ref{fig:overview}). Methods such as the Laplacian Smoothing Loss~\citep{sorkine_laplacian_2004} aim to alleviate this through loss regularization. 
Deployments of CBD in learning applications also craft their own loss regularizers to further manage this distortion~\citep{yifan_neural_2020}.
Such techniques introduce conflicting objectives and optimization methods require careful task-specific parameter tuning to balance competing terms.

In this work we introduce \textit{Structural Affinity}, an algorithm for explicitly regularizing cage-handle motion.
By exploiting the weights of an ``outer cage'' (a cage enclosing the control cage), we obtain an ``affinity'' matrix to propagate deformations between control cage handles.
This maintains the structure prescribed by the outer cage in displacements on the control cage handles, reducing reliance on loss regularization terms.
With $N$ inner-cage handles and $L$ retained off-diagonal affinities, propagation costs $\mathcal{O}(N+L)$ at deformation time and helps avoid the mesh collapse issues of standard CBD.

We show that using CBD-SA enables far easier shape deformation across a range of tasks. In robot morphology co-design, we jointly optimize fingertip shapes and a grasping policy for picking up small objects. In deformation transfer, we train Neural Cages to deform a source shape to match a target. In low-fidelity aerodynamic wing shape optimization, we minimize modeled drag at fixed lift under a geometric bending-energy constraint. By regularising the space of shapes, we enable the more effective use of both gradient-based and non-gradient based optimization for deformation search producing more natural looking shapes.
%========================================================
\section{Background and Prior Work}
\label{sec:bg}

\textbf{Cage-Based Deformation (CBD).} CBD~\citep{gonzalez_hidalgo_cage_2013} encloses a mesh in a coarse control cage whose vertices we call handles. Each mesh vertex is a fixed linear combination of the handles, $\mathbf{v}=\sum_i w_i\mathbf{h}_i$, with weights precomputed at rest, so displacing the handles deforms the mesh smoothly and differentiably. The weighting scheme sets the character of the deformation: mean-value coordinates (MVC)~\citep{ju_mean_2005} give smooth closed-form weights for any closed cage, Green coordinates~\citep{lipman_green_2008} give quasi-conformal deformations, and Somigliana coordinates~\citep{chen_somigliana_2023} allow more elastic behavior.

\textbf{Coupling handle motion.} CBD handles move independently, which can make deformations highly local in dense cages. Modelling tools therefore use distance-based soft selection; \citet{li_learning_2023} map higher-level morphology parameters to cage motion; and direct manipulation~\citep{hsu_direct_1992} solves a least-squares problem for the handle motion that realizes a desired edit.

\textbf{Neural Cages.} Neural networks can deform shapes by predicting new vertex positions~\citep{tan_variational_2018}, a deformation field over 3D space~\citep{wang_3dn_2019}, or cage-handle displacements~\citep{yifan_neural_2020, zhang_kp-red_2024}. Neural Cages~\citep{yifan_neural_2020} predicts both a cage and its handle displacements to match a target shape, enabling deformation transfer. It is trained on ShapeNet~\citep{chang_shapenet_2015} shape pairs with a Chamfer term, an MVC regularizer, and two Laplacian-inspired shape-regularity losses~\citep{wang_3dn_2019, groueix_3d-coded_2018, liu_soft_2019}: $\mathcal{L}_{p2f}$, which penalizes changes in point-to-plane distance over each vertex's 1-ring, and $\mathcal{L}_{normal}$, which penalizes changes in local PCA normals (Appendix~\ref{appx:nc_losses}).

\textbf{Aerodynamic shape optimization.} Aerodynamic design usually parameterizes shapes with B-splines or FFD lattices~\citep{lee_comparison_2017}, often starting from the NASA CRM wing~\citep{noauthor_nasa_2017, chen_aerodynamic_2016, zhang_enhancing_2026}. CBD is rarely used here~\citep{gagnon_two-level_2012, anderson_parametric_2012}, although it can follow the underlying geometry more closely; we hypothesize that the smoother deformations of CBD-SA make it a practical option.

\textbf{Robot morphology co-design.} Co-design jointly optimizes a robot's morphology and its controller for a given task. Differentiable simulators provide gradients with respect to shape~\citep{xu_diffhand_2023, cai_task2morph_2023}, and \citet{li_learning_2023} used CBD to evolve tool shapes such as pushers and scoops for contact-rich manipulation, but propagating gradients through contact remained difficult. Gradient-free optimization of the morphology, with the controller learned alongside, is a common alternative~\citep[e.g.][]{meixner_automated_2019, bjelonic_codesignanymal_2023, tiwary_ifeye_2025}; we use this paradigm, with an RL policy as the controller, to compare CBD and CBD-SA.

%========================================================
\section{Cage-Based Deformation with Structural Affinity}

Structural Affinity couples handle motion without an online solve, using a second, enclosing cage.

In cage-based deformation, a mesh $\mathcal{S}$ is enclosed by a control cage $\mathcal{C}$, which we call the inner cage. With \textit{structural affinity}, $\mathcal{C}$ is itself enclosed by an \textit{outer} cage, $\mathcal{U}$. $\mathcal{S}$ and $\mathcal{C}$
are standard in CBD literature. Every vertex on $\mathcal{S}$ is bound to $\mathcal{C}$ through deformation weights, so moving the vertices of $\mathcal{C}$ induces a deformation of $
\mathcal{S}$.

Similarly, every vertex on $\mathcal{C}$ is bound to $\mathcal{U}$ through a set of generalized barycentric coordinates. In this work, we focus on Mean-Value Coordinates~\citep{ju_mean_2005}.
For every vertex $V_i$ on $\mathcal{C}$, its relationship to the $M$ vertices of $\mathcal{U}$ is defined by a weight vector $W_i=(w_{i1},\ldots,w_{iM})$.

Since $\mathcal{U}$ encloses $\mathcal{C}$, this weight vector describes the position of each inner-cage vertex relative to the larger enclosing structure. We define two vertices on $
\mathcal{C}$ to be structurally related when they have similar weight distributions over $\mathcal{U}$, even if they are not necessarily close in Euclidean distance. We use this observation to
define \textit{structural affinity}: a coefficient $C_{ij}$ describing how strongly motion at inner-cage vertex $V_i$ should propagate to inner-cage vertex $V_j$. We derive these coefficients directly from the outer-cage MVC weights (Section~\ref{sec:sa_derivation}), giving a precomputed relationship between inner-cage vertices which respects the outer cage without requiring an online optimization solve~\citep{jacobson_fast_2012}.

\subsection{Deriving Structural Affinity}
\label{sec:sa_derivation}

We want a single number $C_{ij}$ that says how much $V_j$ should follow when $V_i$ moves. The outer cage already encodes this relationship: moving an outer handle moves every inner-cage vertex that depends on it, by an amount given by that vertex's MVC weight.

Consider first what happens when a single outer-cage handle moves. Each inner-cage vertex is a fixed combination of the outer handles, $V_i=\sum_{m=1}^{M}w_{im}U_m$ with $\sum_{m=1}^{M}w_{im}=1$. Throughout, the outer cage is convex and encloses the inner cage strictly, so every weight is positive~\citep{floater_mean_2005, ju_general_2007}. Moving $U_k$ by $\Delta$ moves $V_i$ by $w_{ik}\Delta$ and $V_j$ by $w_{jk}\Delta$, so $V_j$ follows $V_i$ by the ratio
\begin{equation}
R_{ij}^{k}=\frac{w_{jk}}{w_{ik}}.
\end{equation}

We need to combine the $M$ ratios into one coefficient. A natural choice is to weight each ratio by $U_k$'s share of $V_i$'s motion, $w_{ik}$ (if the whole outer cage translates, $U_k$ contributes a fraction $w_{ik}$ of $V_i$'s motion). Unfortunately, this average is always one: $\sum_{k}w_{ik}R_{ij}^{k}=\sum_{k}w_{jk}=1$. Ratios above one, where $V_j$ is more strongly attached to $U_k$ than $V_i$ is, exactly cancel those below one, so every vertex appears to follow every other vertex completely.

The fix is to let $V_j$ follow $V_i$ without amplifying its motion: we cap each ratio at one before averaging. This gives the structural affinity
\begin{equation}
C_{ij}=\sum_{k=1}^{M}w_{ik}\min\left(1,\frac{w_{jk}}{w_{ik}}\right)=\sum_{k=1}^{M}\min(w_{ik},w_{jk}).
\end{equation}
With the cap, ratios above one no longer compensate for those below, so $C_{ij}$ measures how similarly $V_i$ and $V_j$ are attached to the outer cage. It lies in $[0,1]$, is symmetric, equals one when $i=j$, and is small when the two vertices depend mostly on different outer handles. Appendix~\ref{appx:sa_properties} relates $C_{ij}$ to other known quantities.

Because every weight is positive, $C$ is dense. We sparsify it by keeping the unit diagonal and the largest off-diagonal entries up to a target matrix density $d$, or by removing entries below $\epsilon=10^{-4}$. We do not row-normalize the result, and use this sparse $C$ for deformation propagation.

\subsection{Deformation Propagation}

At deformation time the outer cage is no longer needed: propagation is a single sparse matrix product with the precomputed $C$. 

When a handle $V_i$ is displaced by $z\in\mathbb{R}^3$, every other handle follows in proportion to its affinity,
$V_j'=V_j+\alpha C_{ij}z$ for $j\neq i$, while $V_i$ itself moves by exactly $z$. The affinity strength $\alpha\geq0$ sets how strongly handles follow. Since $C_{ij}\leq1$, for $\alpha\leq1$ no handle moves further than the one displaced.

When several handles move at once, with displacements $D\in\mathbb{R}^{N\times3}$, their contributions add:
\begin{equation}
\Delta V_j=D_j+\alpha\sum_{i\neq j}C_{ij}D_i,
\qquad\text{i.e., since $C_{jj}=1$,}\qquad
\Delta V=AD,\quad A=(1-\alpha)I+\alpha C^{\top}.
\label{eq:sa_update}
\end{equation}
Here $\alpha$ scales only the accumulated propagated displacements.

This formulation does not restrict the inner cage to motions the outer cage can express. In the dense case $C$ is positive semidefinite, so for $0\leq\alpha<1$ every eigenvalue of $A$ is at least $1-\alpha>0$ and $A$ is invertible (Appendix~\ref{appx:sa_properties}): every inner-cage motion remains reachable. Structural Affinity changes which motions are easy to reach, not which are possible. This is the key difference from projecting displacements onto the outer-cage MVC subspace, which removes directions entirely (Appendix~\ref{sec:nc_baselines}).

All outer cages in this work are convex; Appendix~\ref{appx:sa_properties} discusses nonconvex outer cages, whose MVC weights can be negative.

\textbf{Computational Complexity.} The MVC weights and $C$ are computed offline. At deformation time, Eq.~\eqref{eq:sa_update} costs $\mathcal{O}(N+L)$ for $N$ handles and $L$ retained off-diagonal affinities, which is linear in $N$ when each handle keeps a bounded number of neighbours (Appendix~\ref{appx:sa_properties}).

%========================================================
\section{Predicting Cage Deformations with Neural Networks}
\label{sec:nc}

\begin{figure}[!b]
     \centering
     \begin{subfigure}[b]{0.48\linewidth}
         \centering
         \includegraphics[width=\linewidth]{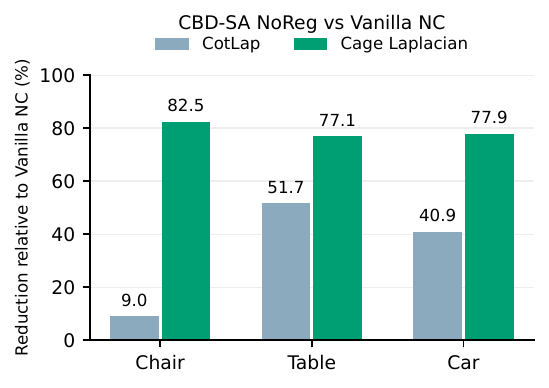}
         \caption{}
         \label{fig_nc:noreg_bars}
     \end{subfigure}
     \hspace{10px}
     \begin{subfigure}[b]{0.48\linewidth}
         \centering
         \includegraphics[width=\linewidth]{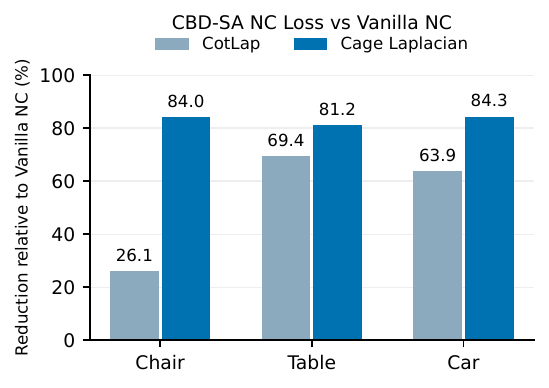}
         \caption{}
         \label{fig_nc:cage_quality_bars}
     \end{subfigure}
     \caption{Cross-category reductions in Cot-Laplacian and effective-cage Laplacian distortion. Percentages compare the six-seed means with Vanilla NC; positive values indicate lower distortion. The left panel isolates Structural Affinity; the right restores the explicit shape-regularity losses.}
     \label{fig_nc:combined_bars}
\end{figure}

We explore the application of CBD-SA to the Neural Cages~\citep{yifan_neural_2020} architecture. For this, we train on the same 3 categories explored by \citet{yifan_neural_2020} from ShapenetV2~\citep{chang_shapenet_2015} - 1) Chair, 2) Table, 3) Car.
Each shape is represented by a point cloud with 5000 points, surface sampled from the base mesh.
We follow the train/test split of Neural Cages~\citep{yifan_neural_2020}, using source–target pairs drawn from within each category.

We report six metrics, all lower-is-better, as mean and sample SD over six seeds: two reconstruction metrics, Density-aware Chamfer Distance (DCD)~\citep{Wu2021DensityawareCD} and Spectrum AUC Difference (SAUCD)~\citep{Luan2024SpectrumAD}, which compare full meshes; the Cot-Laplacian distortion (CotLap) of the deformed mesh; and three cage metrics, measured after affinity propagation: Laplacian distortion, edge distortion, and the fraction of self-intersecting face pairs.

Chamfer Distance (CD), the Neural Cages training target, is reported only in Appendix~\ref{appx:chamfer-shortcomings}: nearest-neighbour CD is a poor measure of mesh alignment, whereas DCD accounts for density mismatch and outliers, and SAUCD agrees with judgments from more than 800 participants~\citep{Luan2024SpectrumAD}.

The baseline comparison is the Vanilla Neural Cages model, trained with the full published loss. We refer to this baseline as \textit{Vanilla NC}. 
We evaluate CBD-SA under 2 loss settings. \textbf{1) NoReg}, a loss profile in which $\mathcal{L}_{p2f}$ and $\mathcal{L}_{normal}$ are removed, to isolate the regularization supplied by the parameterization itself; \textbf{2) Paper Loss}, the full published loss, which measures the additional refinement obtained by restoring the original regularizers.
We refer to these settings as \textit{CBD-SA NoReg} and \textit{CBD-SA NC Loss}.
We use a fixed convex 12-vertex icosahedral outer cage in the normalized category frame, shared across samples and seeds within each category. 
Neural architecture, optimizer hyperparameters, and training schedule are as recommended by \citet{yifan_neural_2020} and are shared between the baseline and the CBD-SA variants.
We report the selected hyperparameters for CBD-SA in Appendix~\ref{appx:hp:nc} and the method by which they were chosen.

\begin{figure}[!b]
     \centering
     \begin{subfigure}[b]{0.25\linewidth}
         \centering
         \includegraphics[width=\linewidth]{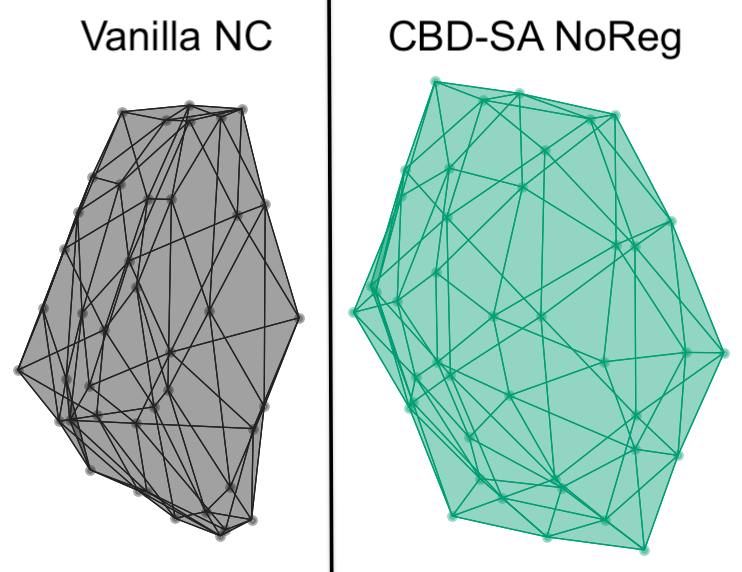}
         \caption{Deformed cages.}
         \label{fig_nc:qualitative_table_cage}
     \end{subfigure}
     \hspace{10px}
     \begin{subfigure}[b]{0.65\linewidth}
         \centering
         \includegraphics[width=\linewidth]{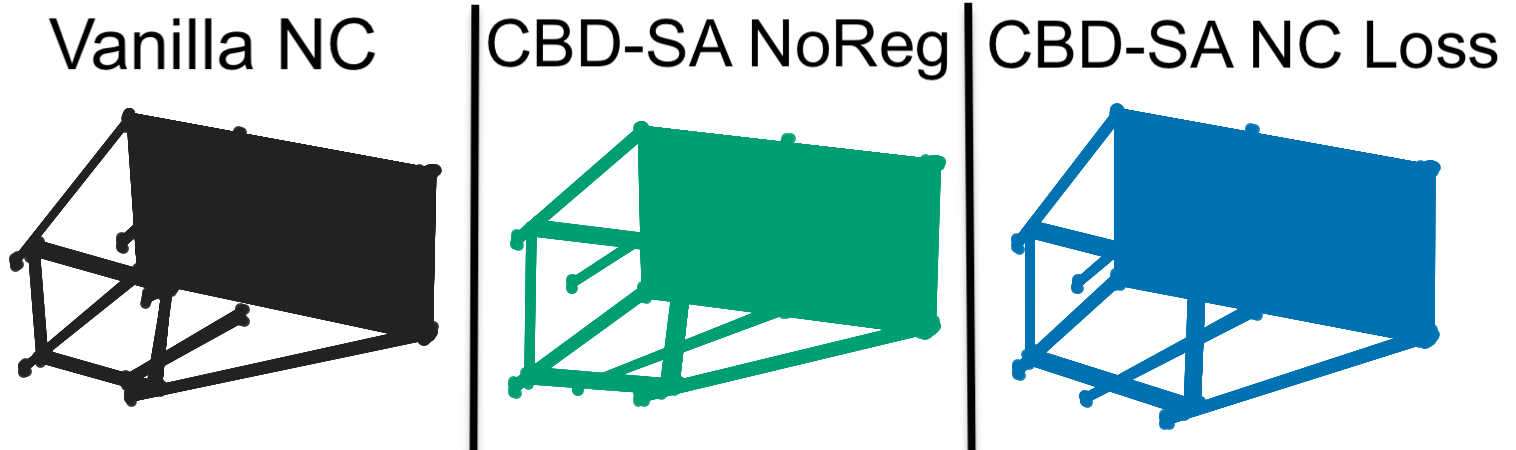}
         \caption{Deformed meshes.}
         \label{fig_nc:qualitative_table_mesh}
     \end{subfigure}
     \caption{Qualitative Table example. (a) Deformed cages for Vanilla NC and CBD-SA NoReg. (b) Deformed meshes for Vanilla NC, CBD-SA NoReg and CBD-SA NC Loss.}
     \label{fig_nc:qualitative_table}
\end{figure}

Most of the cage regularity comes from Structural Affinity itself: without either regularizer, \textit{CBD-SA NoReg} still achieves 93--98\% of the cage-Laplacian improvement and 85--98\% of the cage-edge improvement that \textit{CBD-SA NC Loss} obtains over Vanilla NC (Table~\ref{tab_nc:main_quantitative}). Figure~\ref{fig_nc:combined_bars} shows both loss settings; with the original losses restored, \textit{CBD-SA NC Loss} reduces cage Laplacian distortion by 81--84\% and Cot-Laplacian distortion by 26--69\%. NoReg retains 35--74\% of the Cot-Laplacian gain, so the losses remain complementary for surface smoothness, although computing them adds 42.46\% epoch time and 430 MiB peak GPU memory (Appendix~\ref{appx:compute}). Figure~\ref{fig_nc:qualitative_table_cage} illustrates the cage improvement: the CBD-SA cage stays regular where Vanilla NC tangles.

\begin{table}[!htb]
\centering
\resizebox{\textwidth}{!}{
% Generated by scripts/rebuild_nc_evaluation.py; do not hand-edit numbers.
\begin{tabular}{llcccccc}
\toprule
Category & Method & Cage Lap ($\times10^3$) & Cage Edge & CotLap & Self-Intersect ($\times10^4$) & DCD & SAUCD \\
\midrule
\multirow{3}{*}{Chair} & Vanilla NC & $9.69_{\pm 1.70}$ & $0.285_{\pm 0.018}$ & $10.32_{\pm 3.53}$ & $0.53_{\pm 0.52}$ & \textbf{$0.9100_{\pm 0.0056}$} & $1.5450_{\pm 0.0128}$ \\
 & CBD-SA NoReg & $1.70_{\pm 0.35}$ & $0.110_{\pm 0.007}$ & $9.38_{\pm 4.13}$ & \textbf{$0$} & $0.9153_{\pm 0.0050}$ & \textbf{$1.5393_{\pm 0.0131}$} \\
 & CBD-SA NC Loss & \textbf{$1.55_{\pm 0.17}$} & \textbf{$0.106_{\pm 0.006}$} & \textbf{$7.62_{\pm 2.85}$} & \textbf{$0$} & $0.9144_{\pm 0.0051}$ & $1.5447_{\pm 0.0183}$ \\
\midrule
\multirow{3}{*}{Table} & Vanilla NC & $14.65_{\pm 1.53}$ & $0.346_{\pm 0.024}$ & $145.75_{\pm 188.79}$ & $11.19_{\pm 3.46}$ & \textbf{$0.8716_{\pm 0.0101}$} & \textbf{$1.4316_{\pm 0.0455}$} \\
 & CBD-SA NoReg & $3.36_{\pm 0.54}$ & $0.180_{\pm 0.013}$ & $70.46_{\pm 84.08}$ & $0.29_{\pm 0.50}$ & $0.8790_{\pm 0.0090}$ & $1.4381_{\pm 0.0429}$ \\
 & CBD-SA NC Loss & \textbf{$2.75_{\pm 0.61}$} & \textbf{$0.164_{\pm 0.006}$} & \textbf{$44.64_{\pm 46.89}$} & \textbf{$0$} & $0.8774_{\pm 0.0116}$ & $1.4394_{\pm 0.0318}$ \\
\midrule
\multirow{3}{*}{Car} & Vanilla NC & $0.97_{\pm 0.30}$ & $0.074_{\pm 0.009}$ & $(2.68\pm1.13)\!\times\!10^{-4}$ & \textbf{$0$} & $0.8341_{\pm 0.0057}$ & $1.6765_{\pm 0.0219}$ \\
 & CBD-SA NoReg & $0.21_{\pm 0.02}$ & $0.033_{\pm 0.002}$ & $(1.58\pm0.38)\!\times\!10^{-4}$ & \textbf{$0$} & \textbf{$0.8309_{\pm 0.0056}$} & $1.6853_{\pm 0.0161}$ \\
 & CBD-SA NC Loss & \textbf{$0.15_{\pm 0.02}$} & \textbf{$0.026_{\pm 0.002}$} & \textbf{$(0.97\pm0.30)\!\times\!10^{-4}$} & \textbf{$0$} & $0.8340_{\pm 0.0058}$ & \textbf{$1.6756_{\pm 0.0169}$} \\
\bottomrule
\end{tabular}

}
\vspace{5pt}
\caption{Mean $\pm$ sample SD across six training seeds. Lower is better; the lowest mean per category and metric is bolded. DCD uses the same 5,000 surface samples per prediction as CD; SAUCD uses the corresponding meshes with deterministic reduction when they exceed 10,000 vertices. Chamfer results are retained in Table~\ref{tab_nc:chamfer_compatibility}.}
\label{tab_nc:main_quantitative}
\end{table}

\begin{figure}[!htb]
  \centering
  \includegraphics[width=0.85\linewidth]{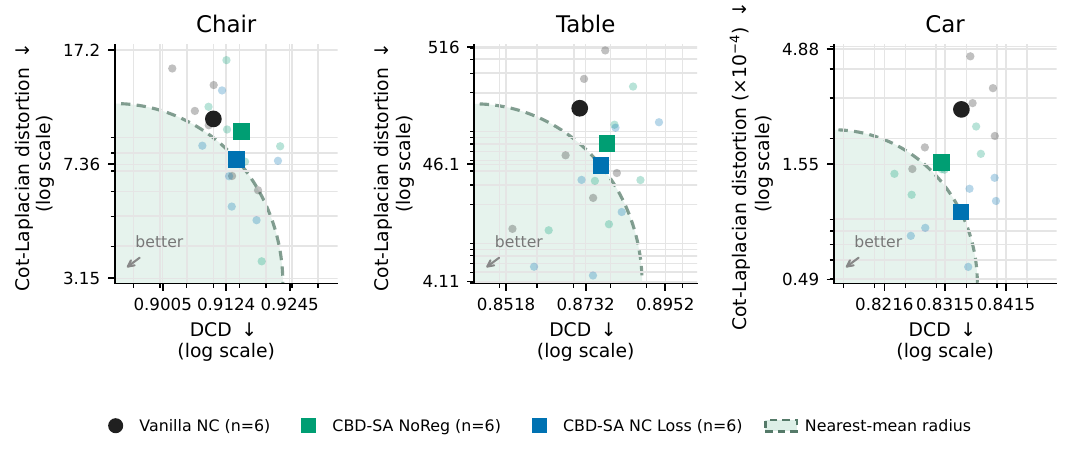}
  \caption{Reconstruction (DCD) against surface distortion (Cot-Laplacian) for Chair, Table and Car, on logarithmic axes; lower is better on both. DCD remains tightly clustered while CBD-SA shifts the means toward lower distortion. The method on the shaded region's boundary, nearest the lower-left corner, has the best joint trade-off. SAUCD shows the same pattern: reconstruction stays tightly clustered while CBD-SA lowers distortion (Figure~\ref{fig_nc:pareto_saucd}, Appendix~\ref{appx:chamfer-shortcomings}).}
  \label{fig_nc:pareto_dcd}
\end{figure}

~\citet{yifan_neural_2020} observe a tradeoff between reconstruction and the cotangent Laplacian of the deformed mesh. Under our complementary reconstruction metrics, this tradeoff is substantially weaker than Chamfer suggests: mean error remains within 1\% of Vanilla NC, and the pooled differences between NoReg and NC Loss are only 0.03\% and 0.05\%. Figure~\ref{fig_nc:pareto_dcd} shows this relationship. The regularity improvement of CBD-SA is therefore much larger than the accompanying fidelity shift, consistent with Figure~\ref{fig_nc:qualitative_table_mesh}.

\textbf{Sensitivity and ablations.} We ablate the affinity strength $\alpha$, the density $d$, and the outer-cage construction in Appendix~\ref{appx:nc_sensitivity}. The outer cage helps as a soft coupling, not as a hard constraint: projecting motion onto the outer-cage MVC subspace does not reproduce the result, and Euclidean Gaussian propagation remains a strong alternative when the outer cage adds little structure (Appendix~\ref{sec:nc_baselines}).
%========================================================
\section{Aerodynamic Shape Optimization}

We apply CBD-SA to wing shape optimization with a low-fidelity vortex-lattice (VLM) variant of the ADODG Case 4.1 CRM drag-minimization benchmark~\citep{lyu_aerodynamic_2015}, using OpenAeroStruct~\citep{Jasa2018a, Chauhan2018b}. The problem keeps the benchmark's structure: minimize drag at fixed lift ($C_L=0.5$), with angle of attack as a trim variable and a bending-energy constraint on the geometry. Because the analysis model is low-fidelity, we compare parameterizations with one another rather than with absolute benchmark optima. We optimize at 3 scales, with 12, 48 and 192 design variables (DVs) for CBD and CBD-SA (Table~\ref{tab:e2-size-levels}, Appendix~\ref{appx:hp:crm}), using SLSQP~\citep{kraft_software_1988, noauthor_numerical_2006, ling_solving_1977} from SciPy~\citep{virtanen_scipy_2020} and IPOPT~\citep{wachter_implementation_2006}, and report the mean and standard deviation across these two deterministic optimizers. Appendix~\ref{appx:aero} gives the flow conditions, mesh, full problem statement, hyperparameters and optimizer settings.

Low drag alone is not enough: a shape can lower $C_D$ numerically through spanwise ripples or local oscillations~\citep{kedward_gradient-limiting_2018}. We therefore compare FFD Baseline, FFD Matched, CBD, and CBD-SA on both the final drag $C_D$ and the deformation roughness $D_R$, the spanwise waviness of the final mesh displacement (Appendix~\ref{appx:hp:crm}).

\begin{figure}[!htb]
\centering
\includegraphics[width=\textwidth]{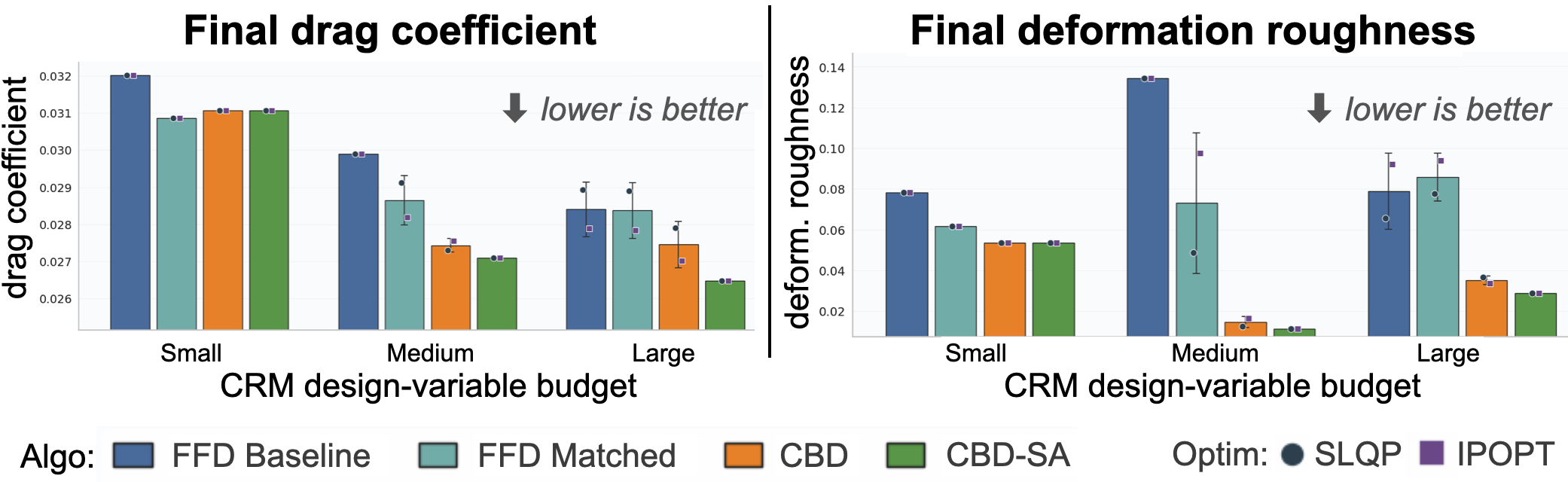}
\caption{Final drag coefficient and deformation roughness of optimized CRM geometry.}
\label{fig:multiopt:cd_and_rough}
\end{figure}

In Figure~\ref{fig:multiopt:cd_and_rough}, we observe that in the small setting, CBD and CBD-SA outperform the FFD Baseline due to an increased number of DVs. With FFD Matched, the methods perform similarly in the small setting. At medium and large, we observe sharp differences. CBD-SA greatly outperforms both FFD settings and vanilla CBD in terms of $C_D$ and $D_R$. We note that this suggests that CBD-SA finds shapes that are aerodynamically better and more geometrically coherent than the evaluated FFD parametrizations under this low-fidelity model. We visualise the final wing geometries for the large scale in Figure~\ref{fig:multiopt:bending_and_shapes} (Appendix~\ref{appx:air_additional}).

With SLSQP and IPOPT, parametrizations satisfy all constraints and convergence is reached. This is important as it ensures that no method achieves lower $C_D$ or $D_R$ by violating a constraint of the problem. Within this original experiment, CBD-SA strictly outperforms on drag and roughness at matched DV count.

\textbf{Physical perturbations.} We additionally test whether the Large CBD-SA result persists across perturbed initial geometries under one fixed optimizer. We evaluate 10 seeds at 5\% and 10\% physical perturbations. In each pair, CBD and CBD-SA start from the same perturbed cage and wing, with the same SLSQP optimizer, 192 active variables, constraints, and 2,000-iteration budget. Each 368-vertex inner cage is enclosed by a 38-vertex convex outer cage, so its MVC weights are positive.

\begin{figure}[H]
\centering
\includegraphics[width=\textwidth]{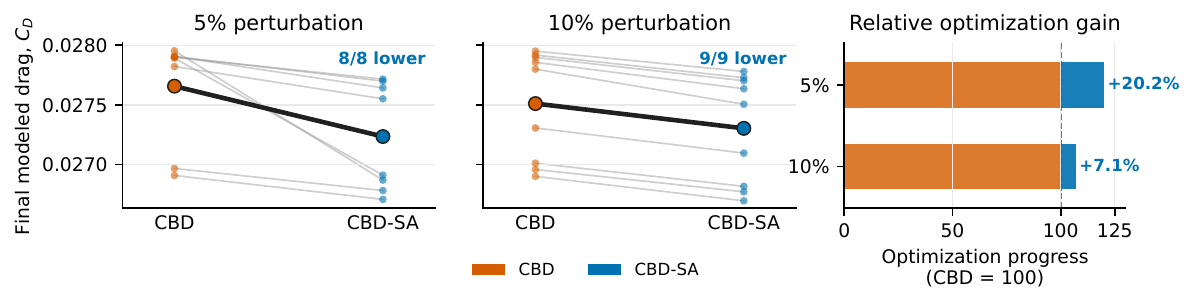}
\caption{Aerodynamic shape-optimization results using the 38-vertex convex outer cages and identical perturbed starts. Left: each thin line is one matched valid run and the thick line connects the mean final drag for CBD and CBD-SA (lower is better). Right: the reduction from the common initialization achieved by CBD is indexed to 100, showing the additional gain from structural affinity.}
\label{fig:crm_refined_paired}
\end{figure}

Structural affinity lowers drag in every run. From identical perturbed starts, CBD-SA reaches lower final modeled drag than CBD in all 8 matched valid pairs at 5\% perturbation and all 9 at 10\% (Figure~\ref{fig:crm_refined_paired}), reducing final $C_D$ by 1.53\% and 0.75\% ($p=0.0078$ and $p=0.0039$, exact paired sign-flip tests). These reductions look small in absolute terms, so we also measure them against the progress CBD itself makes from the common initialization: structural affinity adds a further 20.2\% and 7.14\%. We define this additional optimization gain as $100 \times (\overline{C_D^{\mathrm{CBD}}}-\overline{C_D^{\mathrm{CBD-SA}}})/(\overline{C_D^{\mathrm{initial}}}-\overline{C_D^{\mathrm{CBD}}})$.

\textbf{Outer-cage discretization ablation.} We repeat the study with the coarser convex hulls themselves, which have 19--27 vertices depending on the perturbed cage; the 38-vertex cages above subdivide these hulls without changing the enclosing polytope. CBD-SA again reaches lower modeled drag in every matched valid run (9/9 at both magnitudes), reducing final $C_D$ by 2.06\% and 0.83\% ($p=0.0039$ at both) and providing 27.9\% and 7.9\% additional optimization gain. It also reduces deformation roughness by 18.2\% and 13.0\%, lower in 7/9 pairs at each magnitude (Figure~\ref{fig:crm_primary_paired}; Appendix~\ref{appx:air_additional}). Thus, the lower-drag result does not depend on how finely the convex outer cage is discretized.

%========================================================
\section{Robot Morphology Co-Design}

We optimize finger shapes for the LEAP hand~\citep{shaw_leap_2023} mounted on a PiPER robot arm in a challenging task of picking up small objects from a flat surface. We first pre-train a reinforcement learning (RL) policy~\citep{seo_fasttd3_2025} for simple (pyramidal) fingertip shapes. After this 20-minute pre-training, we run a gradient-free global optimizer~\citep{bennet_nevergrad_2021}. During global optimization (GO) we co-optimize RL policy by conditioning it on the shape parameterization (the cage). This enables simultaneous co-design of the shape and RL policy, nudging RL to extrapolate policies for shapes it has not seen yet.

Our base fingertip mesh has 94 vertices and the structured inner cage has 42 vertices, of which 8 handles positioned around the fingertip are sampled for deformation. Structural affinity uses a convex 8-vertex support box padded around the inner cage.

\textbf{Simulation Experiments.} We quantitatively compare CBD and CBD-SA on the t-nut task using four independent training seeds and an identical 60,000-step co-design budget (approximately eight hours per run). We evaluate each learned policy under two conditions. \emph{Nominal} uses the original softened-contact simulator and evaluation distribution. \emph{Randomized} keeps the learned design and policy fixed while additionally perturbing the object's position ($\pm5$ mm), yaw ($\pm3^\circ$), mass ($0.9$--$1.1\times$), friction ($0.8$--$1.2\times$), contact time constant ($0.004$--$0.012$), and damping ratio ($1.0$--$2.0$). This tests whether performance persists beyond the nominal simulator setting. Each policy is evaluated over 1,000 episodes per condition, using ten evaluation seeds of 100 episodes.

\begin{figure}[t]
    \centering
    \includegraphics[width=\textwidth]{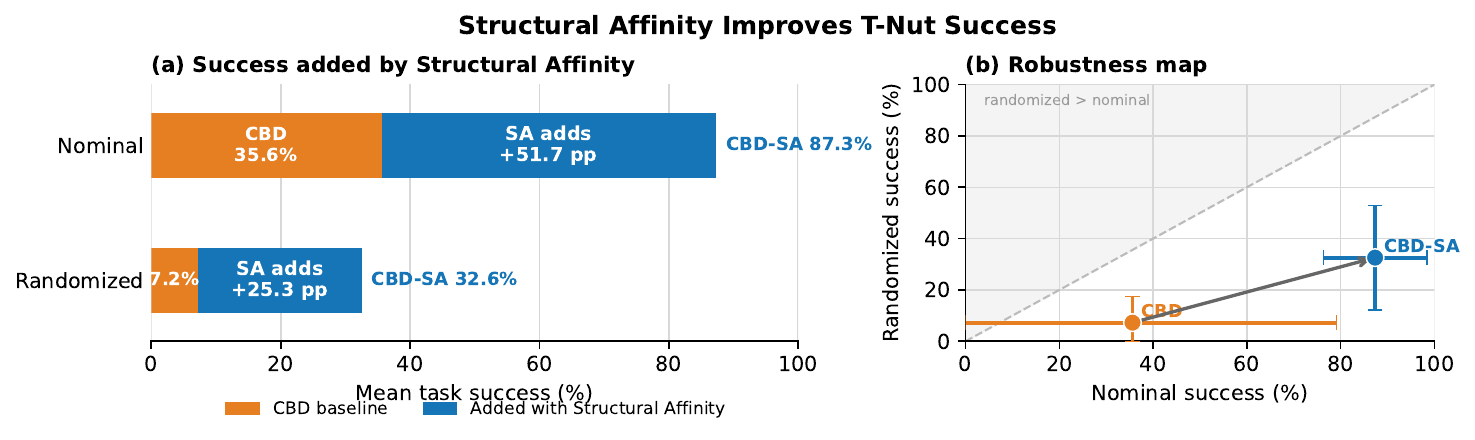}
    \caption{Quantitative t-nut co-design at 60,000 training steps. \emph{Nominal} uses the original softened-contact simulator and evaluation distribution; \emph{Randomized} additionally perturbs object pose, mass, friction, and contact parameters. (a) Mean task success across four independent training seeds; orange shows CBD and the blue segment shows the observed increase with Structural Affinity. (b) Points show means and error bars show sample SD across training seeds. Each learned policy is evaluated over 1,000 episodes per condition, comprising ten evaluation seeds of 100 episodes.}
    \label{fig:robo_tnut_quantitative}
\end{figure}

\begin{figure}[!b]
    \centering
    \includegraphics[width=\textwidth]{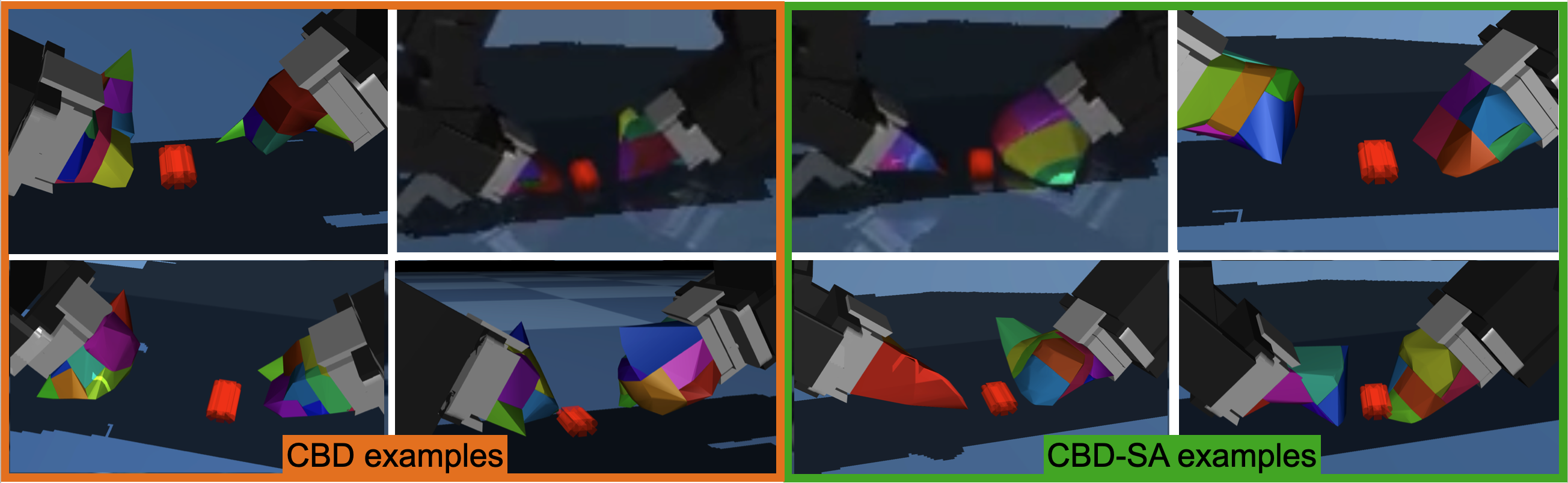}
    \caption{RL+GO optimization with CBD yields spiky fingertips with that exploit simulation by stabbing the object to cause it to stick to fingertips in non-physical ways (a,b). In contrast, using CBD-SA yields fingertips well-suited for pushing the object onto a broad surface provided by one of the fingertips without `stabbing' (c,d).}
    \label{fig:robo_tnut}
    \vspace{-10px}
\end{figure}

\begin{figure}[t]
    \centering
    \includegraphics[width=\textwidth]{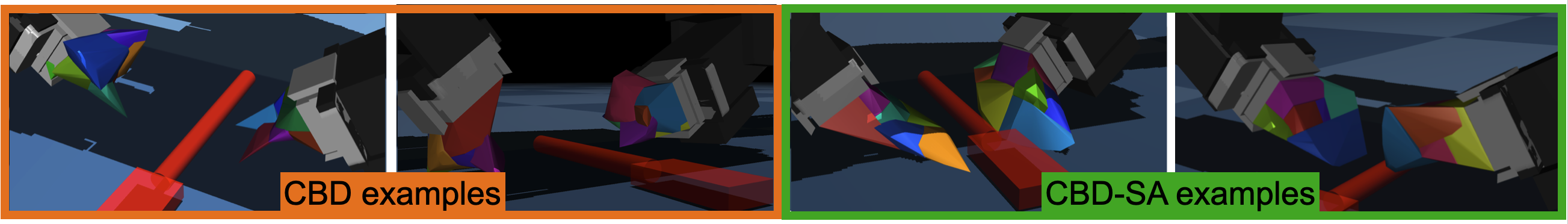}
    \caption{RL+GO optimization with CBD yields jagged fingertips. In contrast, CBD-SA produces fingertips with broader surfaces with physically plausible behavior; Appendix~\ref{app:spatula-lift-comparison} shows the lift frame by frame.}
    \label{fig:robo_spatula}
\end{figure}

Figure~\ref{fig:robo_tnut_quantitative} shows that CBD-SA increases mean t-nut success from $35.60\%\pm43.49\%$ to $87.35\%\pm11.05\%$ under nominal evaluation, and from $7.25\%\pm10.25\%$ to $32.58\%\pm20.35\%$ under randomization (mean $\pm$ sample SD across training seeds). The improvement under both conditions indicates that the benefit is not confined to the chosen simulator parameters. Qualitatively, optimization using vanilla CBD can produce spikier, degenerate shapes that exploit simulation inaccuracies by piercing or knocking objects in non-physical ways. CBD-SA instead produces broader, more manufacturable surfaces that can get underneath small objects and lift them in physically plausible ways, as illustrated for the t-nut in Figure~\ref{fig:robo_tnut}. Figure~\ref{fig:robo_spatula} provides an additional qualitative example on a low-profile spatula with an unbalanced mass distribution.

\begin{figure}[!htb]
    \centering
    \includegraphics[width=\textwidth]{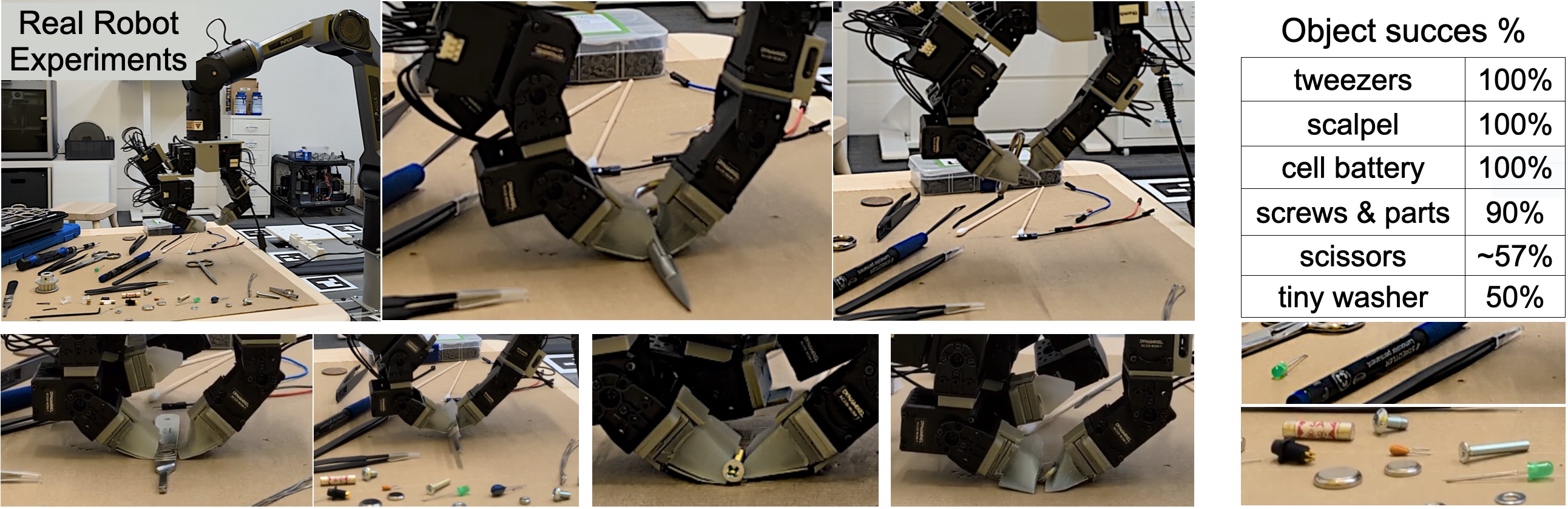}
    \vspace{-15px}
    \caption{Photos of experiments for picking up small, flat, or unbalanced objects from a flat surface.}
    \label{fig:robo_real_results}
\end{figure}

\textbf{Real Robot Experiments.} To validate the capability of CBD-SA to yield shapes that are useful in reality, we manufactured real fingertips using the shapes of the best-performing simulation candidates. For simulation we used MuJoCo with softened contacts (via adjusting solref and solimp parameters), which provided a suitable surrogate for modeling fingertips from a material similar to stiff silicone. We trained RL+GO on three types of objects: tiny objects (nuts \& screws), small but long objects with potentially unballanced center of mass (medical scissors, scalpels, tweezers, small spatulas, long screws with heavy heads), and round flat objects (washers, cell batteries). Then, we 3D printed silicone molds and casted three sets of fingertips. To reduce friction with the table we mounted the fingertips onto 3D printed plastic back plates. Figure~\ref{fig:robo_real_results} shows photos of our real-world setup, photos from several experiments, real objects the robot attempted to pick up, and the relevant success rates. These trials demonstrate physical feasibility of selected CBD-SA designs. We could not manufacture the corresponding CBD fingertips: their thin, spiky features are physically implausible to cast and would not survive contact with the objects.

%========================================================
\FloatBarrier
\section{Conclusion and Future Work}

We have shown that \textit{structural affinity} offers benefits for deformation transfer, aerodynamic shape optimization, and robot morphology co-design. In deformation transfer, Structural Affinity retains 85--98\% of the final cage-level regularity gain without two explicit loss regularizers, while reconstruction fidelity remains within 1\% of Vanilla NC.
Success in these vastly different domains demonstrates broad applicability of the proposed framework.
The key result is that smoother, more regular shapes can be obtained with substantially reduced reliance on tuned loss regularizers. The original losses remain complementary when maximum surface regularity is preferred, but the dominant cage-level effect comes from the parameterization itself. This contributes to effective, compute-efficient search over the shape space. Under the shared low-fidelity aerodynamic model, geometry-informed coupling also produces repeatable modeled-drag improvements across perturbed initial geometries, motivating evaluation under higher-fidelity aerodynamic models.
Outer cage design flexibility allows potential generalization to various structural synergies imposed on the same underlying mesh and inner control-cage. E.g. our method can be applied when the outer cage conforms to that of another geometry for deformation transfer.
The outer cage must still avoid pathological or degenerate constructions.

In this work, we study CBD with mean-value coordinates (MVC), and the affinity matrix formulation is derived from MVC. Ultimately, weighting schemes such as Green Coordinates~\citep{lipman_green_2008}, Bounded Biharmonic Weights~\citep{jacobson_bounded_2011}, and Somigliana Coordinates~\citep{chen_somigliana_2023} could be explored in future work. Deriving affinities under alternative weight formulations is an interesting avenue to explore, as is applying the outer cage binding principle to derive affinity for free-form deformation.

\newpage

\section*{Acknowledgements}

The authors would like to thank Andy Zhou and Yordan Tsvetkov for assistance with proofreading and feedback. 

This project is funded by the Advanced Research + Invention Agency (ARIA) as part of its Robot Dexterity programme under grant number SMRB-P01-P20.

\bibliography{references}

@inproceedings{gonzalez_hidalgo_cage_2013,
    address = {Dordrecht},
    title = {Cage {Based} {Deformations}: {A} {Survey}},
    volume = {7},
    isbn = {978-94-007-5445-4 978-94-007-5446-1},
    shorttitle = {Cage {Based} {Deformations}},
    url = {https://link.springer.com/10.1007/978-94-007-5446-1_3},
    doi = {10.1007/978-94-007-5446-1_3},
    language = {en},
    urldate = {2026-04-15},
    publisher = {Springer Netherlands},
    author = {Nieto, Jesús R. and Susín, Antonio},
    editor = {González Hidalgo, Manuel and Mir Torres, Arnau and Varona Gómez, Javier},
    year = {2013},
    doi = {10.1007/978-94-007-5446-1_3},
    note = {Book Title: Deformation Models
Series Title: Lecture Notes in Computational Vision and Biomechanics},
    pages = {75--99},
}

@inproceedings{xu_diffhand_2023,
  title={An End-to-End Differentiable Framework for Contact-Aware Robot Design},
  author={Xu, Jie and Chen, Tao and Zlokapa, Lara and Foshey, Michael and Matusik, Wojciech and Sueda, Shinjiro and Agrawal, Pulkit},
  booktitle={Robotics: Science \& Systems},
  year={2021}
}

@inproceedings{li_learning_2023,
  title={Learning tool morphology for contact-rich manipulation tasks with differentiable simulation},
  author={Li, Mengxi and Antonova, Rika and Sadigh, Dorsa and Bohg, Jeannette},
  booktitle={IEEE International Conference on Robotics and Automation (ICRA)},
  pages={1859--1865},
  year={2023},
  organization={IEEE}
}

@inproceedings{cai_task2morph_2023,
  author    = {Cai, Yishuai and Yang, Shaowu and Li, Minglong and Chen, Xinglin and Mao, Yunxin and Yi, Xiaodong and Yang, Wenjing},
  title     = {Task2Morph: Differentiable Task-inspired Framework for Contact-Aware Robot Design},
  booktitle = {2023 IEEE/RSJ International Conference on Intelligent Robots and Systems (IROS)},
  year      = {2023},
  pages     = {452--459},
  doi       = {10.1109/IROS55552.2023.10341360}
}

@article{bjelonic_codesignanymal_2023,
  title={Learning-based design and control for quadrupedal robots with parallel-elastic actuators},
  author={Bjelonic, Filip and Lee, Joonho and Arm, Philip and Sako, Dhionis and Tateo, Davide and Peters, Jan and Hutter, Marco},
  journal={IEEE Robotics and Automation Letters},
  volume={8},
  number={3},
  pages={1611--1618},
  year={2023},
  publisher={IEEE}
}

@inproceedings{meixner_automated_2019,
  title={Automated design of simple and robust manipulators for dexterous in-hand manipulation tasks using evolutionary strategies},
  author={Meixner, Andre and Hazard, Christopher and Pollard, Nancy},
  booktitle={2019 IEEE-RAS 19th International Conference on Humanoid Robots (Humanoids)},
  pages={281--288},
  year={2019},
  organization={IEEE}
}

@article{tiwary_ifeye_2025,
  title={What if eye...? Computationally recreating vision evolution},
  author={Tiwary, Kushagra and Young, Aaron and Tasneem, Zaid and Klinghoffer, Tzofi and Dave, Akshat and Poggio, Tomaso and Nilsson, Dan-Eric and Cheung, Brian and Raskar, Ramesh},
  journal={Science Advances},
  volume={11},
  number={51},
  pages={eady2888},
  year={2025},
  publisher={American Association for the Advancement of Science}
}

@misc{yifan_neural_2020,
    title = {Neural {Cages} for {Detail}-{Preserving} {3D} {Deformations}},
    url = {http://arxiv.org/abs/1912.06395},
    doi = {10.48550/arXiv.1912.06395},
    urldate = {2026-04-09},
    publisher = {arXiv},
    author = {Yifan, Wang and Aigerman, Noam and Kim, Vladimir G. and Chaudhuri, Siddhartha and Sorkine-Hornung, Olga},
    month = mar,
    year = {2020},
    note = {arXiv:1912.06395 [cs]},
}

@misc{zhang_kp-red_2024,
    title = {{KP}-{RED}: {Exploiting} {Semantic} {Keypoints} for {Joint} {3D} {Shape} {Retrieval} and {Deformation}},
    shorttitle = {{KP}-{RED}},
    url = {http://arxiv.org/abs/2403.10099},
    doi = {10.48550/arXiv.2403.10099},
    urldate = {2026-04-09},
    publisher = {arXiv},
    author = {Zhang, Ruida and Zhang, Chenyangguang and Di, Yan and Manhardt, Fabian and Liu, Xingyu and Tombari, Federico and Ji, Xiangyang},
    month = dec,
    year = {2024},
    note = {arXiv:2403.10099 [cs]},
}

@article{sederberg_free-form_1986,
    title = {Free-form deformation of solid geometric models},
    volume = {20},
    issn = {0097-8930},
    url = {https://dl.acm.org/doi/10.1145/15886.15903},
    doi = {10.1145/15886.15903},
    language = {en},
    number = {4},
    urldate = {2026-04-15},
    journal = {ACM SIGGRAPH Computer Graphics},
    author = {Sederberg, Thomas W. and Parry, Scott R.},
    month = aug,
    year = {1986},
    pages = {151--160},
}

@inproceedings{sorkine_laplacian_2004,
    address = {New York, NY, USA},
    series = {{SGP} '04},
    title = {Laplacian surface editing},
    isbn = {978-3-905673-13-5},
    url = {https://dl.acm.org/doi/10.1145/1057432.1057456},
    doi = {10.1145/1057432.1057456},
    urldate = {2026-04-15},
    booktitle = {Proceedings of the 2004 {Eurographics}/{ACM} {SIGGRAPH} symposium on {Geometry} processing},
    publisher = {Association for Computing Machinery},
    author = {Sorkine, O. and Cohen-Or, D. and Lipman, Y. and Alexa, M. and Rössl, C. and Seidel, H.-P.},
    month = jul,
    year = {2004},
    pages = {175--184},
}

@inproceedings{shaw_leap_2023,
    title = {{LEAP} {Hand}: {Low}-{Cost}, {Efficient}, and {Anthropomorphic} {Hand} for {Robot} {Learning}},
    isbn = {978-0-9923747-9-2},
    shorttitle = {{LEAP} {Hand}},
    url = {http://www.roboticsproceedings.org/rss19/p089.pdf},
    doi = {10.15607/RSS.2023.XIX.089},
    urldate = {2026-04-15},
    booktitle = {Robotics: {Science} and {Systems} {XIX}},
    publisher = {Robotics: Science and Systems Foundation},
    author = {Shaw, Kenneth and Agarwal, Ananye and Pathak, Deepak},
    month = jul,
    year = {2023},
}

@article{bennet_nevergrad_2021,
    title = {Nevergrad: black-box optimization platform},
    volume = {14},
    shorttitle = {Nevergrad},
    url = {https://dl.acm.org/doi/10.1145/3460310.3460312},
    doi = {10.1145/3460310.3460312},
    number = {1},
    urldate = {2026-04-15},
    journal = {SIGEVOlution},
    author = {Bennet, Pauline and Doerr, Carola and Moreau, Antoine and Rapin, Jeremy and Teytaud, Fabien and Teytaud, Olivier},
    month = apr,
    year = {2021},
    pages = {8--15},
}

@article{hsu_direct_1992,
    title = {Direct manipulation of free-form deformations},
    volume = {26},
    issn = {0097-8930},
    url = {https://dl.acm.org/doi/10.1145/142920.134036},
    doi = {10.1145/142920.134036},
    number = {2},
    urldate = {2026-04-14},
    journal = {SIGGRAPH Comput. Graph.},
    author = {Hsu, William M. and Hughes, John F. and Kaufman, Henry},
    month = jul,
    year = {1992},
    pages = {177--184},
}

@article{ju_mean_2005,
    title = {Mean value coordinates for closed triangular meshes},
    volume = {24},
    issn = {0730-0301, 1557-7368},
    url = {https://dl.acm.org/doi/10.1145/1073204.1073229},
    doi = {10.1145/1073204.1073229},
    language = {en},
    number = {3},
    urldate = {2026-04-15},
    journal = {ACM Transactions on Graphics},
    author = {Ju, Tao and Schaefer, Scott and Warren, Joe},
    month = jul,
    year = {2005},
    pages = {561--566},
}

@article{chen_aerodynamic_2016,
    title = {Aerodynamic {Shape} {Optimization} of {Common} {Research} {Model} {Wing}–{Body}–{Tail} {Configuration}},
    volume = {53},
    issn = {0021-8669},
    url = {https://arc.aiaa.org/doi/10.2514/1.C033328},
    doi = {10.2514/1.C033328},
    number = {1},
    urldate = {2026-05-04},
    journal = {Journal of Aircraft},
    publisher = {American Institute of Aeronautics and Astronautics},
    author = {Chen, Song and Lyu, Zhoujie and Kenway, Gaetan K. W. and Martins, Joaquim R. R. A.},
    month = jan,
    year = {2016},
    pages = {276--293},
}

@article{zhang_enhancing_2026,
    title = {Enhancing surrogate-based aerodynamic shape optimization via dataset-independent generative modelling and dimensionality reduction},
    volume = {20},
    issn = {1994-2060},
    url = {https://doi.org/10.1080/19942060.2026.2629111},
    doi = {10.1080/19942060.2026.2629111},
    number = {1},
    urldate = {2026-04-09},
    journal = {Engineering Applications of Computational Fluid Mechanics},
    publisher = {Taylor \& Francis},
    author = {Zhang, Yang and Han, Zhong-Hua and Zhang, Ke-Shi and Xu, Chen-Zhou and Song, Wen-Ping},
    month = feb,
    year = {2026},
    note = {\_eprint: https://doi.org/10.1080/19942060.2026.2629111},
    pages = {2629111},
}

@misc{noauthor_nasa_2017,
    title = {{NASA} {Common} {Research} {Model} {\textbar} providing data worldwide},
    url = {https://commonresearchmodel.larc.nasa.gov/},
    language = {en-US},
    urldate = {2026-05-04},
    month = aug,
    year = {2017},
}

@article{lee_comparison_2017,
    title = {Comparison of {B}-{Spline} {Surface} and {Free}-{Form} {Deformation} {Geometry} {Control} for {Aerodynamic} {Optimization}},
    volume = {55},
    issn = {0001-1452, 1533-385X},
    url = {https://arc.aiaa.org/doi/10.2514/1.J055102},
    doi = {10.2514/1.J055102},
    language = {en},
    number = {1},
    urldate = {2026-05-04},
    journal = {AIAA Journal},
    author = {Lee, Christopher and Koo, David and Zingg, David W.},
    month = jan,
    year = {2017},
    pages = {228--240},
}

@inproceedings{gagnon_two-level_2012,
    address = {Indianapolis, Indiana},
    title = {Two-{Level} {Free}-{Form} {Deformation} for {High}-{Fidelity} {Aerodynamic} {Shape} {Optimization}},
    isbn = {978-1-60086-930-3},
    url = {https://arc.aiaa.org/doi/10.2514/6.2012-5447},
    doi = {10.2514/6.2012-5447},
    language = {en},
    urldate = {2026-05-04},
    booktitle = {12th {AIAA} {Aviation} {Technology}, {Integration}, and {Operations} ({ATIO}) {Conference} and 14th {AIAA}/{ISSMO} {Multidisciplinary} {Analysis} and {Optimization} {Conference}},
    publisher = {American Institute of Aeronautics and Astronautics},
    author = {Gagnon, Hugo and Zingg, David},
    month = sep,
    year = {2012},
}

@inproceedings{anderson_parametric_2012,
    address = {Nashville, Tennessee},
    title = {Parametric {Deformation} of {Discrete} {Geometry} for {Aerodynamic} {Shape} {Design}},
    isbn = {978-1-60086-936-5},
    url = {https://arc.aiaa.org/doi/10.2514/6.2012-965},
    doi = {10.2514/6.2012-965},
    language = {en},
    urldate = {2026-05-04},
    booktitle = {50th {AIAA} {Aerospace} {Sciences} {Meeting} including the {New} {Horizons} {Forum} and {Aerospace} {Exposition}},
    publisher = {American Institute of Aeronautics and Astronautics},
    author = {Anderson, George and Aftosmis, Michael and Nemec, Marian},
    month = jan,
    year = {2012},
}

@article{tan_variational_2018,
    address = {Salt Lake City, UT},
    title = {Variational {Autoencoders} for {Deforming} {3D} {Mesh} {Models}},
    url = {https://ieeexplore.ieee.org/document/8578710/},
    doi = {10.1109/CVPR.2018.00612},
    urldate = {2026-05-05},
    journal = {2018 IEEE/CVF Conference on Computer Vision and Pattern Recognition},
    publisher = {IEEE},
    author = {Tan, Qingyang and Gao, Lin and Lai, Yu-Kun and Xia, Shihong},
    month = jun,
    year = {2018},
    note = {Conference Name: 2018 IEEE/CVF Conference on Computer Vision and Pattern Recognition (CVPR)
ISBN: 9781538664209},
    pages = {5841--5850},
}

@article{wang_3dn_2019,
    address = {Long Beach, CA, USA},
    title = {{3DN}: {3D} {Deformation} {Network}},
    copyright = {https://doi.org/10.15223/policy-029},
    shorttitle = {{3DN}},
    url = {https://ieeexplore.ieee.org/document/8954215/},
    doi = {10.1109/CVPR.2019.00113},
    urldate = {2026-05-05},
    journal = {2019 IEEE/CVF Conference on Computer Vision and Pattern Recognition (CVPR)},
    publisher = {IEEE},
    author = {Wang, Weiyue and Ceylan, Duygu and Mech, Radomir and Neumann, Ulrich},
    month = jun,
    year = {2019},
    note = {Conference Name: 2019 IEEE/CVF Conference on Computer Vision and Pattern Recognition (CVPR)
ISBN: 9781728132938},
    pages = {1038--1046},
}

@article{chang_shapenet_2015,
    title = {{ShapeNet}: {An} {Information}-{Rich} {3D} {Model} {Repository}},
    shorttitle = {{ShapeNet}},
    url = {https://www.semanticscholar.org/paper/ShapeNet%3A-An-Information-Rich-3D-Model-Repository-Chang-Funkhouser/9b686d76914befea66377ec79c1f9258d70ea7e3},
    urldate = {2026-05-05},
    journal = {ArXiv},
    author = {Chang, Angel X. and Funkhouser, T. and Guibas, L. and Hanrahan, P. and Huang, Qi-Xing and Li, Zimo and Savarese, S. and Savva, M. and Song, Shuran and Su, Hao and Xiao, Jianxiong and Yi, L. and Yu, F.},
    month = dec,
    year = {2015},
}

@inproceedings{groueix_3d-coded_2018,
    address = {Berlin, Heidelberg},
    title = {{3D}-{CODED}: {3D} {Correspondences} by {Deep} {Deformation}},
    isbn = {978-3-030-01215-1},
    shorttitle = {{3D}-{CODED}},
    url = {https://doi.org/10.1007/978-3-030-01216-8_15},
    doi = {10.1007/978-3-030-01216-8_15},
    urldate = {2026-05-05},
    booktitle = {Computer {Vision} – {ECCV} 2018: 15th {European} {Conference}, {Munich}, {Germany}, {September} 8-14, 2018, {Proceedings}, {Part} {II}},
    publisher = {Springer-Verlag},
    author = {Groueix, Thibault and Fisher, Matthew and Kim, Vladimir G. and Russell, Bryan C. and Aubry, Mathieu},
    month = sep,
    year = {2018},
    pages = {235--251},
}

@article{liu_soft_2019,
    address = {Seoul, Korea (South)},
    title = {Soft {Rasterizer}: {A} {Differentiable} {Renderer} for {Image}-{Based} {3D} {Reasoning}},
    copyright = {https://ieeexplore.ieee.org/Xplorehelp/downloads/license-information/IEEE.html},
    shorttitle = {Soft {Rasterizer}},
    url = {https://ieeexplore.ieee.org/document/9008817/},
    doi = {10.1109/ICCV.2019.00780},
    urldate = {2026-05-05},
    journal = {2019 IEEE/CVF International Conference on Computer Vision (ICCV)},
    publisher = {IEEE},
    author = {Liu, Shichen and Chen, Weikai and Li, Tianye and Li, Hao},
    month = oct,
    year = {2019},
    note = {Conference Name: 2019 IEEE/CVF International Conference on Computer Vision (ICCV)
ISBN: 9781728148038},
    pages = {7707--7716},
}

@book{kraft_software_1988,
    series = {Deutsche {Forschungs}- und {Versuchsanstalt} für {Luft}- und {Raumfahrt} {Köln}: {Forschungsbericht}},
    title = {A {Software} {Package} for {Sequential} {Quadratic} {Programming}},
    url = {https://books.google.co.uk/books?id=4rKaGwAACAAJ},
    publisher = {Wiss. Berichtswesen d. DFVLR},
    author = {Kraft, D.},
    year = {1988},
}

@book{noauthor_numerical_2006,
    series = {Springer {Series} in {Operations} {Research} and {Financial} {Engineering}},
    title = {Numerical {Optimization}},
    copyright = {http://www.springer.com/tdm},
    isbn = {978-0-387-30303-1},
    url = {http://link.springer.com/10.1007/978-0-387-40065-5},
    doi = {10.1007/978-0-387-40065-5},
    language = {en},
    urldate = {2026-05-05},
    publisher = {Springer New York},
    year = {2006},
}

@inproceedings{ling_solving_1977,
    title = {Solving {Least} {Squares} {Problems}.},
    volume = {72},
    issn = {01621459},
    url = {https://www.jstor.org/stable/2286501?origin=crossref},
    doi = {10.2307/2286501},
    number = {360},
    urldate = {2026-05-05},
    booktitle = {Journal of the {American} {Statistical} {Association}},
    author = {Ling, Robert F. and Lawson, Charles L. and Hanson, Richard J.},
    month = dec,
    year = {1977},
    note = {Journal Abbreviation: Journal of the American Statistical Association},
    pages = {930},
}

@article{virtanen_scipy_2020,
    title = {{SciPy} 1.0: fundamental algorithms for scientific computing in {Python}},
    volume = {17},
    issn = {1548-7091, 1548-7105},
    shorttitle = {{SciPy} 1.0},
    url = {https://www.nature.com/articles/s41592-019-0686-2},
    doi = {10.1038/s41592-019-0686-2},
    language = {en},
    number = {3},
    urldate = {2026-05-05},
    journal = {Nature Methods},
    author = {Virtanen, Pauli  et al},
    month = mar,
    year = {2020},
    pages = {261--272},
}

@article{wachter_implementation_2006,
    title = {On the implementation of an interior-point filter line-search algorithm for large-scale nonlinear programming},
    volume = {106},
    copyright = {http://www.springer.com/tdm},
    issn = {0025-5610, 1436-4646},
    url = {http://link.springer.com/10.1007/s10107-004-0559-y},
    doi = {10.1007/s10107-004-0559-y},
    language = {en},
    number = {1},
    urldate = {2026-05-05},
    journal = {Mathematical Programming},
    author = {Wächter, Andreas and Biegler, Lorenz T.},
    month = mar,
    year = {2006},
    pages = {25--57},
}

@phdthesis{rago_thesis,
    title        = {Model-Based Derivative-Free Optimization Methods and Software},
    author       = {Ragonneau, T. M.},
    school       = {Department of Applied Mathematics, The Hong Kong Polytechnic University},
    address      = {Hong Kong, China},
    year         = 2022,
    url          = {https://theses.lib.polyu.edu.hk/handle/200/12294},
}

@misc{razh_cobyqa,
    author       = {Ragonneau, T. M. and Zhang, Z.},
    title        = {{COBYQA} {V}ersion 1.1.3},
    year         = 2025,
    url          = {https://www.cobyqa.com},
}

@article{lyu_aerodynamic_2015,
    title = {Aerodynamic {Shape} {Optimization} {Investigations} of the {Common} {Research} {Model} {Wing} {Benchmark}},
    volume = {53},
    issn = {0001-1452, 1533-385X},
    url = {https://mdolab.engin.umich.edu/wiki/aerodynamic-design-optimization-workshop},
    doi = {10.2514/1.J053318},
    language = {en},
    number = {4},
    urldate = {2026-05-05},
    journal = {AIAA Journal},
    author = {Lyu, Zhoujie and Kenway, Gaetan K. W. and Martins, Joaquim R. R. A.},
    month = apr,
    year = {2015},
    pages = {968--985},
}

@article{Jasa2018a,
	Author = {John P. Jasa and John T. Hwang and Joaquim R. R. A. Martins},
	Doi = {10.1007/s00158-018-1912-8},
	Journal = {Structural and Multidisciplinary Optimization},
	Month = {April},
	Number = {4},
	Pages = {1815--1827},
	Publisher = {Springer},
	Title = {Open-source coupled aerostructural optimization using {Python}},
	Volume = {57},
	Year = {2018}}

@inproceedings{Chauhan2018b,
	Author = {Shamsheer S. Chauhan and Joaquim R. R. A. Martins},
	Address = {Lisbon, Portugal},
	Booktitle = {Proceedings of the 6th International Conference on Engineering Optimization, EngOpt 2018},
	Doi = {10.1007/978-3-319-97773-7_38},
	Pages = {418-431},
	Publisher = {Springer},
	Title = {Low-Fidelity Aerostructural Optimization of Aircraft Wings with a Simplified Wingbox Model Using {OpenAeroStruct}},
	Year = {2018}}

@inproceedings{kedward_gradient-limiting_2018,
    address = {Atlanta, Georgia},
    title = {Gradient-{Limiting} {Shape} {Control} for {Efficient} {Aerodynamic} {Optimisation}},
    isbn = {978-1-62410-559-3},
    url = {https://arc.aiaa.org/doi/10.2514/6.2018-3951},
    doi = {10.2514/6.2018-3951},
    language = {en},
    urldate = {2026-05-05},
    booktitle = {2018 {Applied} {Aerodynamics} {Conference}},
    publisher = {American Institute of Aeronautics and Astronautics},
    author = {Kedward, Laurence and Allen, Christian B. and Rendall, T.},
    month = jun,
    year = {2018},
}

@misc{Zhang_2023,
    title        = {{PRIMA: Reference Implementation for Powell's Methods with Modernization and Amelioration}},
    author       = {Zhang, Z.},
    howpublished = {available at http://www.libprima.net, DOI: 10.5281/zenodo.8052654},
    year         = {2023}
}

@inproceedings{alpso,
author = {Li, Changhe and Yang, Shengxiang},
title = {An adaptive learning particle swarm optimizer for function optimization},
year = {2009},
isbn = {9781424429585},
publisher = {IEEE Press},
booktitle = {Proceedings of the Eleventh Conference on Congress on Evolutionary Computation},
pages = {381–388},
numpages = {8},
location = {Trondheim, Norway},
series = {CEC'09}
}

@article{deb_fast_2002,
    title = {A fast and elitist multiobjective genetic algorithm: {NSGA}-{II}},
    volume = {6},
    issn = {1941-0026},
    shorttitle = {A fast and elitist multiobjective genetic algorithm},
    url = {https://ieeexplore.ieee.org/document/996017},
    doi = {10.1109/4235.996017},
    number = {2},
    urldate = {2026-05-06},
    journal = {IEEE Transactions on Evolutionary Computation},
    author = {Deb, K. and Pratap, A. and Agarwal, S. and Meyarivan, T.},
    month = apr,
    year = {2002},
    pages = {182--197},
}

@article{lipman_green_2008,
    title = {Green {Coordinates}},
    volume = {27},
    issn = {0730-0301},
    url = {https://dl.acm.org/doi/10.1145/1360612.1360677},
    doi = {10.1145/1360612.1360677},
    number = {3},
    urldate = {2026-05-06},
    journal = {ACM Trans. Graph.},
    author = {Lipman, Yaron and Levin, David and Cohen-Or, Daniel},
    month = aug,
    year = {2008},
    pages = {1--10},
}

@article{jacobson_bounded_2011,
    title = {Bounded biharmonic weights for real-time deformation},
    volume = {30},
    issn = {0730-0301},
    url = {https://dl.acm.org/doi/10.1145/2010324.1964973},
    doi = {10.1145/2010324.1964973},
    number = {4},
    urldate = {2026-05-06},
    journal = {ACM Trans. Graph.},
    author = {Jacobson, Alec and Baran, Ilya and Popović, Jovan and Sorkine, Olga},
    month = jul,
    year = {2011},
    pages = {78:1--78:8},
}

@inproceedings{chen_somigliana_2023,
    address = {New York, NY, USA},
    series = {{SIGGRAPH} '23},
    title = {Somigliana {Coordinates}: an elasticity-derived approach for cage deformation},
    isbn = {979-8-4007-0159-7},
    shorttitle = {Somigliana {Coordinates}},
    url = {https://dl.acm.org/doi/10.1145/3588432.3591519},
    doi = {10.1145/3588432.3591519},
    urldate = {2026-05-06},
    booktitle = {{ACM} {SIGGRAPH} 2023 {Conference} {Proceedings}},
    publisher = {Association for Computing Machinery},
    author = {Chen, Jiong and De Goes, Fernando and Desbrun, Mathieu},
    month = jul,
    year = {2023},
    pages = {1--8},
}

@article{jacobson_fast_2012,
    title = {Fast automatic skinning transformations},
    volume = {31},
    issn = {0730-0301},
    url = {https://dl.acm.org/doi/10.1145/2185520.2185573},
    doi = {10.1145/2185520.2185573},
    number = {4},
    urldate = {2026-04-15},
    journal = {ACM Trans. Graph.},
    author = {Jacobson, Alec and Baran, Ilya and Kavan, Ladislav and Popović, Jovan and Sorkine, Olga},
    month = jul,
    year = {2012},
    pages = {77:1--77:10},
}

@article{seo_fasttd3_2025,
    title = {{FastTD3}: {Simple}, {Fast}, and {Capable} {Reinforcement} {Learning} for {Humanoid} {Control}},
    copyright = {Creative Commons Attribution 4.0 International},
    shorttitle = {{FastTD3}},
    url = {https://arxiv.org/abs/2505.22642},
    doi = {10.48550/ARXIV.2505.22642},
    urldate = {2026-05-07},
    publisher = {arXiv},
    author = {Seo, Younggyo and Sferrazza, Carmelo and Geng, Haoran and Nauman, Michal and Yin, Zhao-Heng and Abbeel, Pieter},
    year = {2025},
    note = {Version Number: 3},
}

@article{Luan2024SpectrumAD,
  title={Spectrum AUC Difference (SAUCD): Human-Aligned 3D Shape Evaluation},
  author={Tianyu Luan and Zhong Li and Lele Chen and Xuan Gong and Li-Chang Chen and Yi Xu and Junsong Yuan},
  journal={2024 IEEE/CVF Conference on Computer Vision and Pattern Recognition (CVPR)},
  year={2024},
  pages={20155-20164},
  url={https://api.semanticscholar.org/CorpusID:268248874}
}

@article{Wu2021DensityawareCD,
  title={Density-aware Chamfer Distance as a Comprehensive Metric for Point Cloud Completion},
  author={Tong Wu and Liang Pan and Junzhe Zhang and Tai Wang and Ziwei Liu and Dahua Lin},
  journal={ArXiv},
  year={2021},
  volume={abs/2111.12702},
  url={https://api.semanticscholar.org/CorpusID:244527380}
}

@article{Zeng2022CD2F3,
  title={CD2: Fine-grained 3D Mesh Reconstruction with Twice Chamfer Distance},
  author={Rongfei Zeng and Mai Su and Ruiyun Yu and Xingwei Wang},
  journal={ACM Transactions on Multimedia Computing, Communications and Applications},
  year={2022},
  volume={19},
  pages={1 - 25},
  url={https://api.semanticscholar.org/CorpusID:249240454}
}

@inproceedings{
sharifipour2025apml,
title={{APML}: Adaptive Probabilistic Matching Loss for Robust 3D Point Cloud Reconstruction},
author={Sasan Sharifipour and Constantino Alvarez Casado and Mohammad Sabokrou and Miguel Bordallo Lopez},
booktitle={The Thirty-ninth Annual Conference on Neural Information Processing Systems},
year={2025},
url={https://openreview.net/forum?id=PqAGmIg0az}
}

@inproceedings{Jin2019DRKFSAD,
  title={DR-KFS: A Differentiable Visual Similarity Metric for 3D Shape Reconstruction},
  author={Jiongchao Jin and Akshay Gadi Patil and Zhang Xiong and Hao Zhang},
  booktitle={European Conference on Computer Vision},
  year={2019},
  url={https://api.semanticscholar.org/CorpusID:216411241}
}

@inproceedings{barla_histogram_2003,
  title={Histogram Intersection Kernel for Image Classification},
  author={Barla, Annalisa and Odone, Francesca and Verri, Alessandro},
  booktitle={Proceedings of the International Conference on Image Processing},
  year={2003},
  pages={513--516}
}

@article{floater_mean_2005,
    title = {Mean value coordinates in {3D}},
    volume = {22},
    number = {7},
    journal = {Computer Aided Geometric Design},
    author = {Floater, Michael S. and K{\'o}s, G{\'e}za and Reimers, Martin},
    year = {2005},
    pages = {623--631},
    doi = {10.1016/j.cagd.2005.06.004},
}

@article{ju_general_2007,
    title = {A general geometric construction of coordinates in a convex simplicial polytope},
    volume = {24},
    number = {3},
    journal = {Computer Aided Geometric Design},
    author = {Ju, Tao and Liepa, Peter and Warren, Joe},
    year = {2007},
    pages = {161--178},
    doi = {10.1016/j.cagd.2006.12.001},
}
\bibliographystyle{thestyle}

\appendix
%========================================================
\clearpage
\appendix
\section*{Appendix}

\FloatBarrier
\section{Structural Affinity Details}
\label{appx:sa_properties}

\FloatBarrier
\subsection{Properties}
\paragraph{Equivalent forms.}
Because both MVC rows sum to one, $\min(a,b)=\frac{1}{2}(a+b-|a-b|)$ gives
\begin{equation}
C_{ij}=1-\frac{1}{2}\sum_{k=1}^{M}|w_{ik}-w_{jk}|,
\end{equation}
so $C_{ij}$ is one minus the total-variation distance between the two coordinate distributions. Equivalently, $C_{ij}$ is the largest mass the two rows share,
\begin{equation}
C_{ij}=\max_{\substack{q\in\mathbb{R}_{\geq0}^{M}\\q\leq W_i,\;q\leq W_j}}\sum_{k=1}^{M}q_k,
\end{equation}
attained at $q_k=\min(w_{ik},w_{jk})$. This is the histogram-intersection kernel~\citep{barla_histogram_2003} applied to MVC rows; what is new here is using it to couple inner-cage handles through outer-cage geometry.

\paragraph{Invertibility of the propagation operator.}
For $a,b\in[0,1]$, $\min(a,b)=\int_0^1\mathbf{1}[a\geq t]\mathbf{1}[b\geq t]\,dt$. Hence for every $x\in\mathbb{R}^{N}$,
\begin{equation}
x^\top Cx=\sum_{k=1}^{M}\int_0^1\left(\sum_{i=1}^{N}x_i\mathbf{1}[w_{ik}\geq t]\right)^2dt\geq0,
\end{equation}
so the dense $C$ is positive semidefinite. Since the dense $C$ is also symmetric, $A$ in Eq.~\eqref{eq:sa_update} equals $(1-\alpha)I+\alpha C$, and every eigenvalue of $A$ is at least $1-\alpha$, and $A$ is invertible for $0\leq\alpha<1$. Sparsification can break positive semidefiniteness, so this guarantee does not carry over automatically; the sparse operator is still invertible whenever $\alpha\lVert C-I\rVert_2<1$.

\paragraph{Nonconvex outer cages.}
For a nonconvex outer cage, MVC weights can be negative even for enclosed points. The identity $C_{ij}=1-\frac{1}{2}\lVert W_i-W_j\rVert_1$ still holds, so $C$ remains symmetric with $C_{ii}=1$ and $C_{ij}\leq1$, but it can become negative: $W_i=(1,0)$ and $W_j=(1+t,-t)$ give $C_{ij}=1-t<0$ for $t>1$. The capped-ratio derivation and the shared-mass form no longer apply. If nonnegative propagation is needed, the coefficients can be clipped, $\widehat{C}_{ij}=\max(0,\min(1,C_{ij}))$, which keeps symmetry and the unit diagonal but does not guarantee a valid deformation. A convex outer cage avoids the issue.

\paragraph{Limitations of the propagation update.}
Bounded pairwise affinities do not guarantee an intersection-free cage when contributions from several handles accumulate. The update is also not normalized, so it does not preserve a common translation: if every handle moves by $z$, handle $j$ moves by $(1+\alpha\sum_{i\neq j}C_{ij})z$. In our applications, global pose is fixed or handled separately.

\FloatBarrier
\subsection{Computational Complexity}
The expensive quantities are computed offline: the MVC weights $w_{im}$ and the affinity coefficients $C_{ij}$. At deformation time, $C$ is fixed, so propagation only requires applying the stored coefficients to the user's inner-cage displacement.

Let $N$ be the number of inner-cage vertices, and let $L$ be the number of non-zero off-diagonal affinity coefficients kept after thresholding. For a single selected inner-cage handle $V_i$, the update costs $T_i = \mathcal{O}(1+\ell_i)$, where $\ell_i$ is the number of other vertices with non-zero affinity to $V_i$. If the implementation writes all $N$ inner-cage vertex positions into an output array, the cost is $\mathcal{O}(N+\ell_i)$.

For multiple simultaneously edited inner-cage handles, let $D \in \mathbb{R}^{N\times3}$ store the input displacement at each inner-cage vertex. The resulting displacement is given by Eq.~\eqref{eq:sa_update}, with
$T = \mathcal{O}(N+L)$.
If each inner-cage vertex keeps only $s$ non-zero off-diagonal affinity coefficients on average, then $L=sN$ and $T=\mathcal{O}(N+sN)=\mathcal{O}(N)$ for fixed $s$. The dense worst case keeps all pairwise affinities, so $L=N(N-1)$ and propagation becomes $\mathcal{O}(N^2)$. In practice, removing weak affinities reduces the online cost; linear scaling requires the average number of retained neighbors to remain bounded as $N$ grows, rather than holding the affinity density fixed.

\FloatBarrier
\section{Neural Cages}

\FloatBarrier
\subsection{Shape-Regularity Losses}
\label{appx:nc_losses}
\citet{yifan_neural_2020} introduce and use 2 primary shape regularity loss terms inspired by Laplacian regularization~\citep{wang_3dn_2019, groueix_3d-coded_2018, liu_soft_2019}, $\mathcal{L}_{p2f}$ and $\mathcal{L}_{normal}$. $\mathcal{L}_{p2f}$ fits a PCA plane to the 1-ring neighborhood of vertices on the mesh and computes point-to-plane distance ($d_i$) and penalises the change in this distance post deformation ($d_i'$). $\mathcal{L}_{normal}$ quantifies the angular difference of PCA normals before ($n_i$) and after ($n_i'$) deformation, which are reported to improve the quality of the deformation~\citep{yifan_neural_2020}. These losses are defined by the following equations where $V_{s}$ is the set of vertices of the source shape:
$
\mathcal{L}_{p2f} = \frac{1}{|V_{s}|} \sum_{i=1}^{|V_s|} ||d_i - d_i'||^2 \; \; \; \text{and} \; \; \; \mathcal{L}_{normal} = \frac{1}{|V_{s}|} \sum_{i=1}^{|V_{s}|} \left( 1 - n_{i}^{T} n_{i}' \right)
$

\FloatBarrier
\subsection{Evaluation Metrics and Chamfer Distance}
\label{appx:chamfer-shortcomings}
\paragraph{Metric definitions.}
We evaluate our method using six metrics, where lower values indicate better performance across all measures: \textbf{(1) Density-aware Chamfer Distance (DCD)~\citep{Wu2021DensityawareCD}}, measuring density-aware point-set alignment between the deformed source and target; \textbf{(2) Spectrum AUC Difference (SAUCD)~\citep{Luan2024SpectrumAD}}, measuring perceptual alignment between their full meshes; \textbf{(3) Cot-Laplacian Distortion (CotLap)}, assessing the preservation of local surface details on the deformed mesh; \textbf{(4) Cage Laplacian Distortion}, capturing the smoothness regularity of the cage after affinity propagation; \textbf{(5) Cage Edge Distortion}, measuring edge stretching relative to the rest pose; and \textbf{(6) Cage Self-Intersections}, the fraction of nonadjacent cage-face pairs that intersect (ideally zero). All values are reported as mean and sample standard deviation over six seeds.

Chamfer has important limitations in the setting of structural affinity, which targets cohesive deformation. Chamfer is a pointwise cost rather than a measure of structural alignment, so the magnitude of a small change cannot by itself establish a consequential change in geometric quality. Mesh regularity and cage validity are therefore important complementary indicators of deformation quality.

This limitation is independently documented. ~\citet{Luan2024SpectrumAD} compare 3D shape metrics against annotations from more than 800 participants. A particularly illustrative example appears in Figure~1, where a slight pose error obtains Chamfer \(1.35\), while an over-smoothed distortion obtains Chamfer \(0.40\). Further, the authors report that Chamfer has low overall correlation with human judgments: Pearson \(0.097\), Spearman \(0.079\), and Kendall \(0.080\). ~\citet{Jin2019DRKFSAD} present a chair example in which modifying all four legs produces a larger Chamfer distance than removing one leg.

Further analysis from~\citet{sharifipour2025apml} show that the nearest-neighbor assignments induced by Chamfer distance can produce many-to-one correspondences, point congestion, and poor coverage. Work by~\citet{Wu2021DensityawareCD} further indicates that Chamfer can be insensitive to mismatched local density, overly influenced by outliers, and inconsistent as an evaluation metric. ~\citet{Zeng2022CD2F3} additionally document that Chamfer-based mesh deformation can produce vertex clustering and twisted or intersecting faces because point-level nearest-neighbor matching ignores mesh edges and global shape structure. The authors show that Chamfer can therefore favor meshes with worse validity.

The broader literature establishes that Chamfer captures one useful but narrow property: local pointwise proximity. Small variations in point-sampling strategies and minor point offsets can accumulate, and Chamfer does not reliably encode perceptual importance, part correspondence, mesh connectivity, or structural validity---all of which are central to coherent, nondegenerate deformation and to our setting. There is little correspondence between a percentage change in Chamfer and a consequential change in quality at the scales observed in our experiments.

We do not claim that Chamfer is invalid, but it alone cannot determine the magnitude of perceptual or structural degradation. We therefore report DCD and SAUCD, which mitigate complementary shortcomings of Chamfer Distance, alongside regularity metrics. Across all categories and both CBD-SA settings, these two fidelity metrics remain within 1\% of Vanilla NC, whereas CBD-SA NC Loss reduces Cot-Laplacian distortion by 26\%--69\% and cage Laplacian distortion by 81\%--84\%. The stronger reconstruction metrics thus indicate a much smaller fidelity shift than the corresponding gain in regularity.

\FloatBarrier
\subsection{Hyperparameters}
\label{appx:hp:nc}
For Neural Cages, we choose hyperparameters by grid search.

For these experiments, we obtained separate hyperparameters for each category which were maintained across 6 evaluation seeds. These are detailed in Table~\ref{tab:nc_hyperparameters}. The search was conducted on the CBD-SA NoReg setting, and hyperparameters were transferred to CBD-SA NC Loss.

\begin{table}[!htb]
\centering
\begin{tabular}{lcc}
\toprule
Category & Affinity $\alpha$ & Affinity Density $d$ \\
\midrule
Chair & 0.3 & 0.4 \\
Table & 0.2 & 0.4 \\
Car   & 0.4 & 0.4 \\
\bottomrule
\end{tabular}
\vspace{5pt}
\caption{Selected hyperparameters per category: $\alpha$ controls affinity propagation strength; $d$ controls the target graph density. Each category uses a fixed convex 12-vertex icosahedral support in its normalized category frame, shared across samples and seeds. Affinity weights use a min-weight aggregation scheme.}
\label{tab:nc_hyperparameters}
\end{table}

For Neural Cages we reused the optimizer settings outlined by the original work~\citep{yifan_neural_2020}.

\FloatBarrier
\subsection{Neural Cages Sensitivity and Outer-Cage Construction}
\label{appx:nc_sensitivity}

The canonical outer cage is a convex subdivision-0 icosphere computed once in the shared normalized category frame. Chair, Table, and Car use 12 vertices and 20 faces; the category-specific scale or radii are listed in Table~\ref{tab:nc_outer_cages}. No mesh decimation or convex fallback is used. The learned 42-vertex inner cage, source mesh, and target mesh remain sample dependent.

\begin{table}[!htb]
\centering
\begin{tabular}{lccc}
\toprule
Category & Training shapes used for bounds & Outer cage & Scale/radii \\
\midrule
Chair & 6,440 & 12V/20F & scale 2.25, margin 0.15 \\
Table & 8,015 & 12V/20F & $(2.40,2.15,2.22)$ \\
Car & 3,339 & 12V/20F & $(2.20,2.15,2.40)$ \\
\bottomrule
\end{tabular}
\vspace{5pt}
\caption{Canonical Neural Cages outer supports. Each cage is fixed within a category and shared across samples, seeds, and loss profiles.}
\label{tab:nc_outer_cages}
\end{table}

The outer cage acts as a structural support rather than a tightly fitted deformation cage. In practice, it should (i) enclose the inner cage throughout the evaluated data, (ii) remain nondegenerate, and (iii) represent the broad geometric relationships that should be propagated between handles. A simple convex support is sufficient for the applications studied here. For normalized category-level data, as in Neural Cages, one fixed support can be shared across all samples and seeds. For geometries whose proportions vary, as in aerodynamic optimization, we instead construct a convex hull from each padded inner cage and verify enclosure. A more informative outer-cage geometry may encode richer relationships, but is not required to obtain the regularization effect demonstrated here.

We evaluated 22 Chair hyperparameter configurations spanning affinity strength $\alpha$ and target density $d$. The parameters have complementary roles: $\alpha$ scales only the displacement propagated to related handles, leaving each handle's prescribed displacement unchanged, while $d$ determines the retained affinity connections. Consequently, small $\alpha$ or sparse $d$ weakens propagation and approaches ordinary CBD. At the opposite extreme, strong, dense coupling makes many handles move collectively, approaching the behavior of a substantially coarser effective cage and reducing local deformation freedom. Between these limits, $\alpha\in[0.1,0.45]$ and $d\in[0.2,0.7]$ have provided a useful practical range across our Neural Cages experiments. These ranges are practical recommendations rather than strict admissible bounds: 21 of the 22 evaluated configurations improve cage Laplacian distortion over Vanilla NC, and 15 eliminate cage self-intersections. The strongest cage regularity within a $+5\%$ mean-Chamfer compatibility constraint occurs at $\alpha=0.6,d=0.2$, outside the usual $\alpha$ range, with cage Laplacian distortion $0.001194$ and $+3.89\%$ Chamfer.

To separate robustness from parameter retuning, we hold $\alpha=0.3$ and $d=0.4$ fixed while changing only the Chair outer cage. Table~\ref{tab:nc_outer_sensitivity} compares the canonical 12-vertex support with a coarser enclosure and a support perturbed by approximately 20\% of the canonical cage's mean radius. The coarser and perturbed-cage experiments use three seeds, while the canonical result uses six. Both alternatives retain low cage-Laplacian distortion, showing that the result does not depend on one precisely positioned support. This should not be interpreted as invariance to arbitrary cages: a degenerate support or one encoding unsuitable structural relationships can propagate unsuitable motion.

\begin{table}[!htb]
\centering
\begin{tabular}{lccc}
\toprule
Outer cage & Seeds & Cage Laplacian $\downarrow$ & Chamfer $\downarrow$ \\
\midrule
Canonical 12-vertex support & 6 & $0.001548\pm0.000169$ & $0.02869\pm0.00168$ \\
Coarser enclosing support & 3 & $0.001061\pm0.000301$ & $0.03066\pm0.00097$ \\
20\%-perturbed support & 3 & $0.001336\pm0.000067$ & $0.02940\pm0.00132$ \\
\bottomrule
\end{tabular}
\vspace{5pt}
\caption{Chair outer-cage sensitivity, reported as mean $\pm$ sample SD. Chamfer is included only for compatibility with the original Neural Cages evaluation.}
\label{tab:nc_outer_sensitivity}
\end{table}

\FloatBarrier
\subsection{Ablations: Direct Projection and Gaussian Propagation}
\label{sec:nc_baselines}
We compare Structural Affinity against two simpler ways of coupling handle motion. \textbf{Direct projection} uses the same outer-cage MVC matrix $W\in\mathbb{R}^{N\times M}$ as CBD-SA, but as a hard constraint: on each forward pass, the predicted inner-handle displacements $D\in\mathbb{R}^{N\times3}$ are projected into the column space of $W$,
\begin{equation}
D_{\mathrm{proj}}=P_WD,\qquad P_W=WW^{+},
\end{equation}
where $W^{+}$ is the Moore--Penrose pseudoinverse. This is the most direct way to use the outer cage: it tests whether outer-cage coordinates help as a hard constraint rather than as a soft coupling. \textbf{Gaussian propagation} instead couples inner handles by their Euclidean distance, without using the outer cage: it replaces $C$ in Eq.~\eqref{eq:sa_update} with $C^{\mathrm{G}}_{ij}=\exp(-\lVert V_i-V_j\rVert^2/\sigma^2)$, sparsified with the same density rule as CBD-SA, and selects $\alpha$, $d$ and $\sigma$ by grid search. We choose it because distance-based soft selection is the standard way modelling tools couple handles (Section~\ref{sec:bg}), so it tests whether the outer cage adds anything beyond Euclidean proximity. Both baselines keep the Neural Cages predictor's 42 inner-handle outputs; only the forward-pass displacement map differs.

Table~\ref{tab:direct_projection} compares the direct outer-MVC projection with CBD-SA under the same paper-loss setting. Across Chair, Table, and Car, CBD-SA has lower effective-cage Laplacian and edge distortion, with the largest separation on Table. Reconstruction remains close: in the paired evaluation pooled over the three categories, CBD-SA's DCD is $0.20\%$ higher and its valid-pair SAUCD is $0.63\%$ lower than direct projection. The hard subspace therefore does not explain the observed combination of cage regularity and reconstruction fidelity. It remains a useful constraint, but SA preserves full inner-handle control while biasing the way those controls interact.

\begin{table}[!htb]
\centering
\caption{Direct outer-MVC projection and CBD-SA with the published Neural Cages losses. Values are means across six matched seeds; lower is better. Cage metrics are measured on the effective cage used for deformation, after each method's forward-pass map.}
\label{tab:direct_projection}
\begin{tabular}{llcc}
\toprule
Category & Method & Effective cage Laplacian ($\times10^3$) & Effective cage edge \\
\midrule
Chair & Direct projection & 8.92 & 0.254 \\
      & CBD-SA & 5.18 & 0.232 \\
Table & Direct projection & 20.79 & 0.350 \\
      & CBD-SA & 7.26 & 0.285 \\
Car   & Direct projection & 0.644 & 0.0716 \\
      & CBD-SA & 0.630 & 0.0667 \\
\bottomrule
\end{tabular}
\end{table}

The Gaussian comparison tests a different alternative: coupling by Euclidean proximity, trained with the NoReg loss profile so that only the propagation rules differ. With the canonical outer cage, CBD-SA is more regular at similar reconstruction (Section~\ref{appx:nc_gaussian}). Under the canonical outer cage, the two methods select many of the same handle relationships, so we do not interpret their regularity difference as universal kernel superiority. Their NoReg comparison nevertheless shows that this overlap does not make the two propagation rules identical. When the outer cage carries more category geometry, the selected relationships diverge further; in the evaluated coarse-cage setting CBD-SA lowers pooled effective cage-edge distortion relative to Gaussian by $4.48\%$, with category Chamfer differences below $5\%$. Other Gaussian settings can be preferable on some metrics. The supported distinction is therefore an additional source of geometric information and a useful operating regime, not a claim that min-weight affinity always dominates Euclidean smoothing. Appendix~\ref{appx:informative_cage} gives the cage construction and comparison protocol.

Taken together, these ablations show that the outer cage matters as a soft coupling, not as a hard constraint: projecting onto outer-cage MVC coordinates does not reproduce the result, and Euclidean propagation remains a strong alternative when the outer cage carries little extra structure. The benefit of Structural Affinity therefore depends on the structure encoded by the outer cage; we do not claim that min-weight overlap is optimal for every cage or task.

\FloatBarrier
\subsubsection{Gaussian NoReg Comparison}
\label{appx:nc_gaussian}
To compare the regularization supplied by the propagation rules themselves, we remove $\mathcal{L}_{p2f}$ and $\mathcal{L}_{normal}$ from both CBD-SA and the Euclidean Gaussian baseline. Gaussian propagation weights inner-cage handle pairs by their Euclidean separation, whereas CBD-SA uses their outer-cage MVC relationships. The CBD-SA comparison uses the canonical convex category-level support.

\begin{figure}[!htb]
  \centering
  \includegraphics[width=\linewidth]{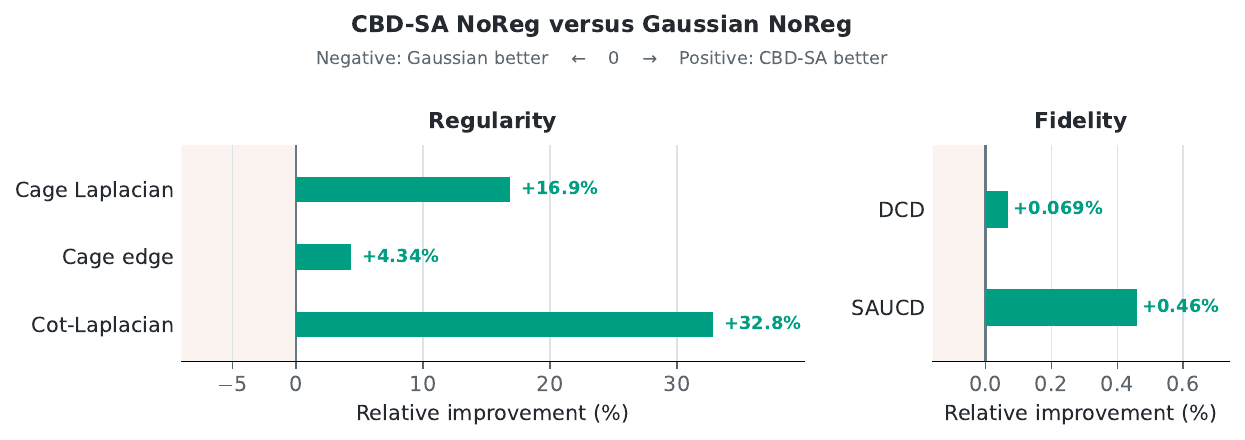}
  \caption{CBD-SA NoReg compared with Gaussian NoReg using the canonical outer cage. Positive values favor CBD-SA; the panels use separate horizontal scales to show the smaller reconstruction differences. Relative improvements are reported for Cot-Laplacian error, predicted-cage Laplacian and edge distortion, DCD, and SAUCD. Lower is better for every metric.}
  \label{fig:nc_gaussian_noreg}
\end{figure}

Figure~\ref{fig:nc_gaussian_noreg} shows that the observed differences are concentrated in regularity: the largest relative improvement is in Cot-Laplacian error ($32.8\%$), followed by predicted-cage Laplacian ($16.9\%$) and edge distortion ($4.34\%$). Pooled DCD and SAUCD improve by only $0.069\%$ and $0.46\%$, respectively. Thus, the additional regularity is not accompanied by a corresponding loss of reconstruction fidelity under these metrics. The comparison also isolates a difference between the propagation rules that can be obscured when both methods are trained with the explicit regularization losses. Even with the generic canonical support, outer-cage MVC relationships provide a distinct coupling of handles from Euclidean distance alone, and the observed pattern is consistent with a stronger regularizing prior. These are descriptive comparisons across the evaluated categories, rather than a claim that every category or seed improves by the same amount.

\FloatBarrier
\subsubsection{Baseline Details}
\label{appx:informative_cage}
The direct-projection comparison uses the same 42-output Neural Cages predictor, outer-coordinate matrix $W$, published losses, six training seeds, and evaluation pairs as CBD-SA. Projection is applied to the predicted inner-handle displacement at every forward pass; it changes the effective displacement subspace without reducing the number of network outputs. With $N=42$, $M=12$ and $\operatorname{rank}(W)=12$, projection leaves 36 rather than 126 geometric displacement degrees of freedom, whereas the dense CBD-SA operator is full rank for $\alpha<1$. Table~\ref{tab:direct_projection} reports effective-cage rather than unpropagated predicted-cage regularity.

The canonical Neural Cages outer support is a fixed convex 12-vertex icosahedral cage in the normalized category frame. For the informative-cage comparison, a coarse support is generated by inflating and coarsening the inner cage so that its MVC coordinates reflect category structure more closely than the fixed canonical support. The support was not manually sculpted for individual examples or categories. The restricted Table study selected the coarse-cage CBD-SA setting $\alpha=0.2$, $d=0.4$ and Gaussian setting $\alpha=0.1$, $d=0.6$, $\sigma=0.5$; the construction and settings were then frozen for evaluation on Chair, Table, and Car with six matched seeds. Gaussian propagation itself does not use the outer cage. Its comparison support is used only to measure graph overlap at equal density.

With the canonical support, CBD-SA and Gaussian share $87.5$--$94.2\%$ of selected handle relationships across categories; with the informative coarse support they share $82.2$--$83.7\%$. In the frozen coarse-cage comparison, CBD-SA has $4.48\%$ lower pooled effective cage-edge distortion than Gaussian; the category differences are $6.04\%$ on Chair, $2.08\%$ on Table, and $5.18\%$ on Car. The corresponding Chamfer increases are $4.45\%$, $3.83\%$, and $0.82\%$. These observations show where outer-coordinate geometry can distinguish SA from a distance-only kernel, while leaving open whether other kernels built from the same coordinates could yield similar behavior. The canonical and informative comparisons are distinct settings, and neither implies that CBD-SA dominates every Gaussian configuration.

\FloatBarrier
\subsection{Additional Results}
\begin{figure}[!htb]
  \centering
  \includegraphics[width=0.85\linewidth]{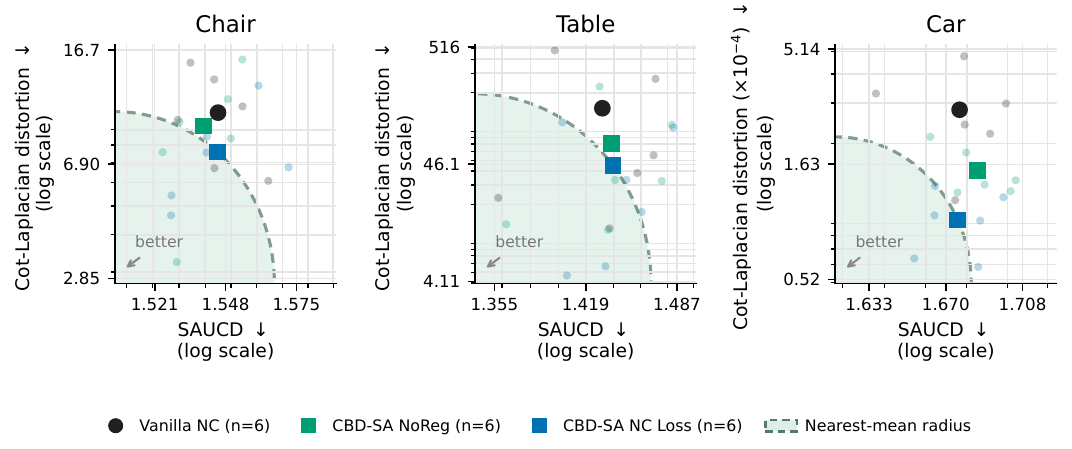}
  \caption{Reconstruction (SAUCD) against surface distortion (Cot-Laplacian) for Chair, Table and Car, on logarithmic axes; lower is better on both. SAUCD remains tightly clustered while CBD-SA shifts the means toward lower distortion. The shaded region extends from the lower-left corner to the nearest method mean; the method on its boundary has the best joint trade-off.}
  \label{fig_nc:pareto_saucd}
\end{figure}

\begin{table}[!htb]
\centering
% Generated by scripts/rebuild_nc_evaluation.py.
\begin{tabular}{llccc}
\toprule
Category & Method & CD ($\times10^2$) & DCD & SAUCD \\
\midrule
\multirow{3}{*}{Chair} & Vanilla NC & \textbf{$2.658_{\pm 0.157}$} & \textbf{$0.9100_{\pm 0.0056}$} & $1.5450_{\pm 0.0128}$ \\
 & CBD-SA NoReg & $2.876_{\pm 0.195}$ & $0.9153_{\pm 0.0050}$ & \textbf{$1.5393_{\pm 0.0131}$} \\
 & CBD-SA NC Loss & $2.869_{\pm 0.168}$ & $0.9144_{\pm 0.0051}$ & $1.5447_{\pm 0.0183}$ \\
\midrule
\multirow{3}{*}{Table} & Vanilla NC & \textbf{$3.166_{\pm 0.219}$} & \textbf{$0.8716_{\pm 0.0101}$} & \textbf{$1.4316_{\pm 0.0455}$} \\
 & CBD-SA NoReg & $3.301_{\pm 0.268}$ & $0.8790_{\pm 0.0090}$ & $1.4381_{\pm 0.0429}$ \\
 & CBD-SA NC Loss & $3.516_{\pm 0.394}$ & $0.8774_{\pm 0.0116}$ & $1.4394_{\pm 0.0318}$ \\
\midrule
\multirow{3}{*}{Car} & Vanilla NC & $0.733_{\pm 0.137}$ & $0.8341_{\pm 0.0057}$ & $1.6765_{\pm 0.0219}$ \\
 & CBD-SA NoReg & \textbf{$0.702_{\pm 0.131}$} & \textbf{$0.8309_{\pm 0.0056}$} & $1.6853_{\pm 0.0161}$ \\
 & CBD-SA NC Loss & $0.733_{\pm 0.146}$ & $0.8340_{\pm 0.0058}$ & \textbf{$1.6756_{\pm 0.0169}$} \\
\bottomrule
\end{tabular}

\caption{Reconstruction metrics from the same evaluation as Table~\ref{tab_nc:main_quantitative}, retaining Chamfer Distance (CD) for comparison with the original training objective. Mean $\pm$ sample SD across six training seeds. Lower is better; the lowest mean is bolded. CD is multiplied by $10^2$ for readability.}
\label{tab_nc:chamfer_compatibility}
\end{table}

\FloatBarrier
\section{Aerodynamic Shape Optimization}
\label{appx:aero}

\FloatBarrier
\subsection{Problem Setup}
\label{appx:hp:crm}
The setup keeps the structure of the ADODG Case 4.1 benchmark but is not intended to reproduce its RANS benchmark values. Instead, it uses the OAS CRM lifting-surface setup, with viscous drag correction enabled and wave drag disabled, on a fixed coarse aerodynamic mesh. Additional optimizers, such as COBYLA~\citep{Zhang_2023}, COBYQA~\citep{razh_cobyqa, rago_thesis}, NSGA2~\citep{deb_fast_2002}, ALPSO~\citep{alpso} proved computationally intractable for this problem.

The CRM is a configuration prescribed by NASA which is representative of a widebody commercial transport aircraft. It is specifically designed for cruise Mach number $M_{\infty} = 0.85$ and design lift coefficient $C_L = 0.5$; following the OAS CRM example, we analyse it at $M_{\infty} = 0.84$ (Table~\ref{tab:e2-crm-flow-surface}). In our setting, the CRM wing is created through the OAS CRM mesh generator. Table~\ref{tab:e2-crm-flow-surface} shows the single aerodynamic condition under which our wing is optimized. The CRM mesh is generated with \(3\) chordwise vertices and \(17\) spanwise vertices for the full wing. Because symmetry is enabled, the analyzed mesh is the half-wing mesh with \(N_x=3\) chordwise stations and \(N_y=9\) spanwise stations.

In the drag minimization problem, we optimize for both the angle of attack, $\alpha$, and the geometry of the wing, $\mathbf{x}$. The solver's control over $\mathbf{x}$ is specified by the parametrization. In the case of FFD, this corresponds to positions of the lattice control-points. For CBD and CBD-SA, this corresponds to positions of the control/inner cage handles. $\alpha$ is initialized to 2.75, and $\mathbf{x}$ is initialized as the base CRM geometry with a twist, to give a non-standard starting point to optimization. The overall optimization problem we solve can then be written as constrained minimization: $\min_{\alpha,\,\mathbf{x}} C_D(\alpha,\mathbf{x}) \quad \mathrm{s.t.} \quad
       C_L(\alpha,\mathbf{x}) = 0.5,
       E_{\mathrm{bend}}(\mathbf{x}) \le 0.05,
       C_D(\alpha,\mathbf{x}) \ge 10^{-5},
       -10^\circ \le \alpha \le 10^\circ, 
       \mathbf{x}_{\min} \le \mathbf{x} \le \mathbf{x}_{\max}$.

For each optimized wing, let \(d_{ij}=z_{ij}-z^0_{ij}\) be the vertical mesh displacement relative to the baseline. We average this over chordwise stations and measure spanwise waviness by the following formula. A larger \(D_R\) indicates a more oscillatory spanwise deformation: $\bar d_j = \frac{1}{N_x}\sum_{i=1}^{N_x} d_{ij}, \;\text{for} \; j=1,\ldots,N_y; \ D_R =
  \left(
  \sum_{j=2}^{N_y-1}
  \left(
  \bar d_{j+1} - 2\bar d_j + \bar d_{j-1}
  \right)^2
  \right)^{1/2}$.

\begin{table}[!htb]
  \centering
  \small
  \setlength{\tabcolsep}{2.5pt}
  \begin{tabular}{lcccc}
  \toprule
  Resolution & \makecell{FFD Base\\Grid} & \makecell{FFD Base\\DVs} & \makecell{FFD Match\\DVs} & \makecell{CBD-SA\\DVs} \\
  \midrule
  Small  & $2 \times 3$  & $6$ & $12$ & $12$  \\
  Medium & $4 \times 6$  & $24$ & $48$ & $48$  \\
  Large  & $8 \times 12$ & $96$ & $192$ & $192$ \\
  \bottomrule
  \end{tabular}
  \caption{Geometry-control resolutions used in the CRM parameterization study. The aerodynamic mesh is fixed for all cases; the shape-control space varies.}
  \label{tab:e2-size-levels}
\end{table}

In our experiments, we use parameters outlined in Table~\ref{tab:e2-crm-flow-surface}, which correspond to the setup from the OAS CRM example\footnote{Available at https://mdolab-openaerostruct.readthedocs-hosted.com/en/latest/aero\_walkthrough.html}.

\begin{table}[!htb]
  \centering
  \begin{tabular}{ll}
  \hline
  \textbf{Quantity} & \textbf{Value} \\
  \hline
  Geometry & NASA Common Research Model (CRM) wing \\
  Symmetry & Half-wing symmetry \\
  Aerodynamic mesh & $N_x = 3,\; N_y = 17$ \\
  Baseline twist & CRM twist baked into the baseline mesh \\
  Mach number & $M_\infty = 0.84$ \\
  Freestream density & $\rho_\infty = 0.38~\mathrm{kg\,m^{-3}}$ \\
  Freestream velocity & $V_\infty = 248.136~\mathrm{m\,s^{-1}}$ \\
  Reynolds number per unit length & $\mathrm{Re}/m = 1.0 \times 10^{6}~\mathrm{m^{-1}}$ \\
  Initial angle of attack & $\alpha_0 = 2.75^\circ$ \\
  Sideslip angle & $\beta = 0^\circ$ \\
  Target lift coefficient & $C_L^\star = 0.5$ \\
  Reference area convention & Wetted area \\
  Baseline lift offset & $C_{L,0} = 0$ \\
  Baseline drag offset & $C_{D,0} = 0.015$ \\
  Laminar fraction parameter & $k_{\mathrm{lam}} = 0.05$ \\
  Thickness-to-chord control point & $(t/c) = 0.15$ \\
  Maximum thickness location & $x_{t,\max}/c = 0.303$ \\
  Viscous drag correction & Enabled \\
  Wave drag correction & Disabled \\
  \hline
  \end{tabular}
  \vspace{5pt}
    \caption{Flow, aerodynamic model, and surface settings used for the CRM experiments.}
  \label{tab:e2-crm-flow-surface}
\end{table}

\FloatBarrier
\subsection{Hyperparameters and Perturbation Protocol}
For our experiments, we obtained separate hyperparameters per scale, but maintained them across optimizers. These are detailed in Table~\ref{tab:crm_hyperparameters}.

\begin{table}[!htb]
\centering
\begin{tabular}{lcc}
\toprule
Scale & Affinity $\alpha$ & Affinity Density $d$ \\
\midrule
Small & 0.1 & 0.05 \\
Medium & 0.7 & 0.15 \\
Large   & 0.7 & 0.1 \\
\bottomrule
\end{tabular}
\vspace{5pt}
\caption{Selected hyperparameters per scale. $\alpha$ controls affinity propagation strength; $d$ controls the target graph density. The outer cage is computed once per scale. Affinity weights use a min-weight aggregation scheme.}
\label{tab:crm_hyperparameters}
\end{table}

For the perturbed-geometry experiments, exploratory seeds 0--9 were used to select $\alpha_{\mathrm{SA}}=0.05$ and target density $d=0.025$. We then froze this hyperparameter pair for held-out seeds 10--19 at 5\% and 10\% perturbation. Within each seed, CBD and CBD-SA began from the same perturbed physical reference cage and wing, with zero displacement and identical aerodynamic models, constraints, bounds, derivatives, SLSQP settings, and 2,000-iteration budgets. The Large cage exposes 192 active vertical-displacement variables.

For each perturbed 368-vertex inner cage, we apply a fixed radial padding and take its convex hull, followed by a half-space enclosure check. The resulting held-out supports have 19--27 vertices and nonnegative raw MVC weights. For the support-discretization ablation, hull facets are subdivided to produce a 38-vertex, 72-face cage without changing the enclosing convex polytope. The additional optimization gain is computed as
\[
100\times\frac{\overline{C_D^{\mathrm{CBD}}}-\overline{C_D^{\mathrm{CBD-SA}}}}
{\overline{C_D^{\mathrm{initial}}}-\overline{C_D^{\mathrm{CBD}}}},
\]
using matched valid runs. It is the additional reduction obtained by structural affinity relative to the reduction already achieved by CBD, rather than a percentage reduction in final drag.

\FloatBarrier
\subsection{Optimizer Settings}
\label{appx:opt}
For aerodynamic shape optimization we used the optimizer settings specified in Table~\ref{tab:crm_optimizer_settings}.

 \begin{table}[!htb]
    \centering
    \small
    \begin{tabularx}{\linewidth}{@{}l l l X@{}}
      \toprule
      Optimizer & Driver & Main limits & Additional settings \\
      \midrule
      SLSQP
      & SciPy
      & \begin{tabular}[t]{@{}l@{}}
          \texttt{tol = 1e-9} \\
          \texttt{maxiter = 200} small/medium \\
          \texttt{maxiter = 2000} large
        \end{tabular}
      & \begin{tabular}[t]{@{}l@{}}
          \texttt{objective\_scaler = 1e4} \\
          \texttt{debug\_print = false} \\
          \texttt{print\_results = false} \\
          \texttt{opt\_settings = \{\}}
        \end{tabular} \\
      \addlinespace

      IPOPT
      & pyOptSparse
      & \begin{tabular}[t]{@{}l@{}}
          \texttt{tol = 1e-6} \\
          \texttt{maxiter = 3000} \\
          \texttt{max\_iter = 3000}
        \end{tabular}
      & \begin{tabular}[t]{@{}l@{}}
          \texttt{objective\_scaler = 1e4} \\
          \texttt{gradient\_method = openmdao} \\
          \texttt{acceptable\_tol = 1e-5} \\
          \texttt{acceptable\_iter = 15} \\
          \texttt{hessian\_approximation = limited-memory} \\
          \texttt{limited\_memory\_max\_history = 1000} \\
          \texttt{mu\_strategy = monotone} \\
          \texttt{mu\_init = 1e-5} \\
          \texttt{nlp\_scaling\_method = none} \\
          \texttt{start\_with\_resto = yes} \\
          \texttt{max\_resto\_iter = 200}
        \end{tabular} \\
      \bottomrule
    \end{tabularx}
    \vspace{5pt}
    \caption{Optimizer settings used for the CRM multi-optimizer comparison.}
    \label{tab:crm_optimizer_settings}
  \end{table}

\FloatBarrier
\subsection{Additional Results}
\label{appx:air_additional}
\begin{figure}[!htb]
\centering
\includegraphics[width=0.7\textwidth]{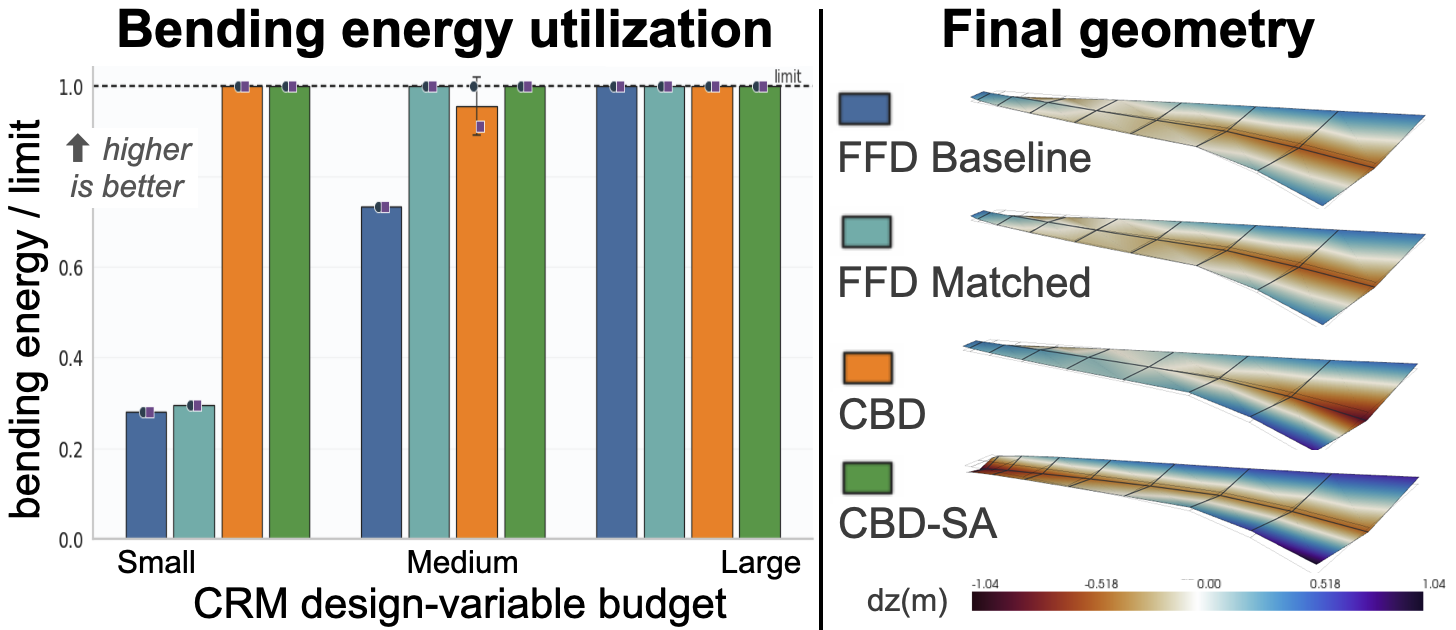}
\caption{Final bending energy of optimized CRM geometry.}
\label{fig:multiopt:bending_and_shapes}
\end{figure}

\begin{figure}[!htb]
\centering
\includegraphics[width=0.62\textwidth]
{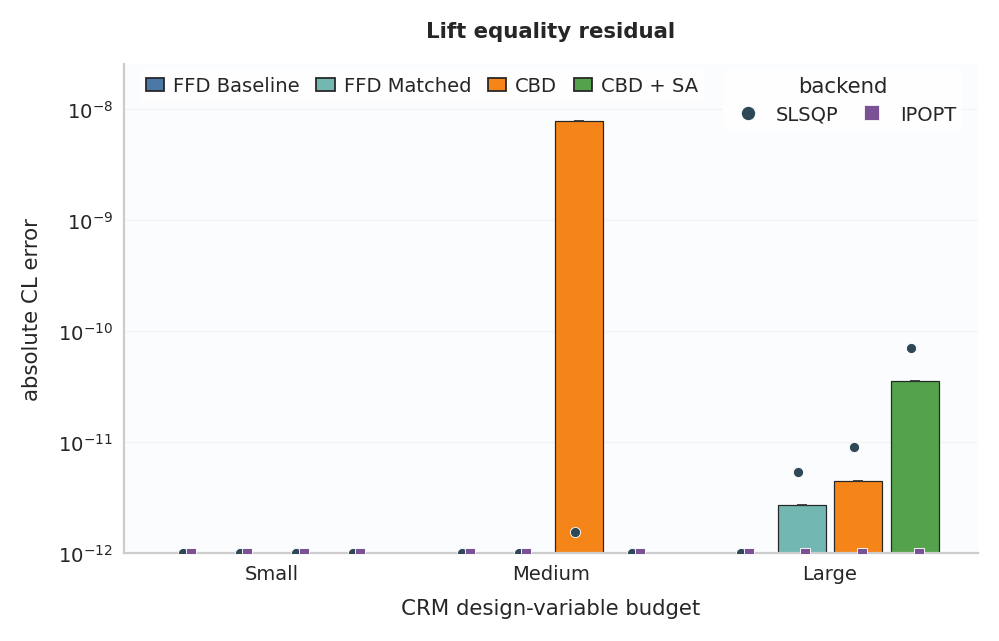}
\caption{Final CL residual of optimized CRM geometry.}
\label{fig:multiopt:cl}
\end{figure}
Figure~\ref{fig:multiopt:cl} shows the final $C_L$ residual achieved by each method post optimization. The maximum residual achieved is $10^{-8}$ by CBD at the medium scale. This indicates that all optimizers on all methods and scales converged successfully, as $10^{-8}$ is well within the typical optimizer tolerance for these problems.

Table~\ref{tab:crm_perturbed_all} reports the two convex-support perturbation studies separately. The 19--27-vertex study uses the directly constructed convex hulls; the 38-vertex study tests sensitivity to support discretization. Differences and confidence intervals are computed from paired final $C_D$ values, with negative values favoring CBD-SA.

\begin{figure}[!htb]
\centering
\includegraphics[width=\textwidth]{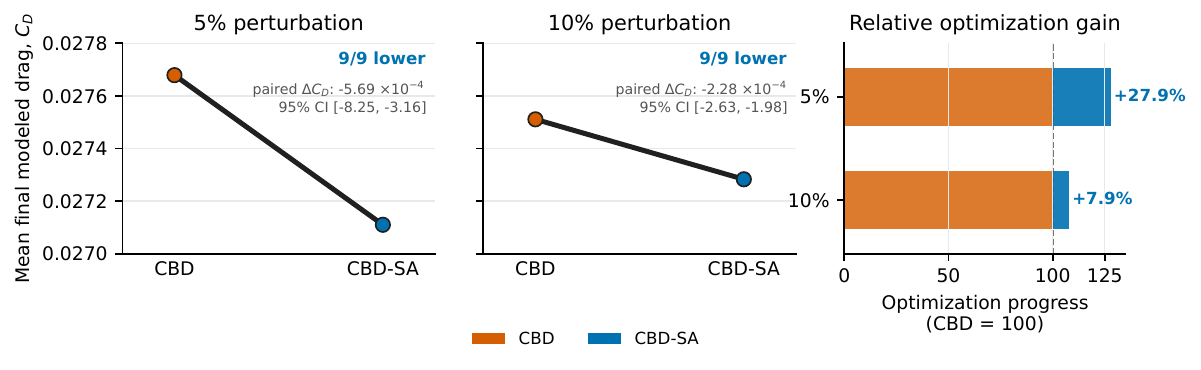}
\caption{Outer-cage discretization ablation using the 19--27-vertex convex hulls and identical perturbed starts. Left: mean final modeled drag for CBD and CBD-SA, together with the paired effect interval and the number of matched runs favoring CBD-SA (lower is better). Right: the reduction from the common initialization achieved by CBD is indexed to 100, showing the additional optimization gain from structural affinity.}
\label{fig:crm_primary_paired}
\end{figure}

\begin{table}[!htb]
\centering
\small
\setlength{\tabcolsep}{3.5pt}
\begin{tabular}{llcccc}
\toprule
Outer cage & Perturb. & Matched & Mean $\Delta C_D$ & 95\% CI & Reduction \\
\midrule
19--27 vertices & 5\%  & 9 & $-0.000569$ & $[-0.000825,-0.000316]$ & 2.06\% \\
19--27 vertices & 10\% & 9 & $-0.000228$ & $[-0.000263,-0.000198]$ & 0.83\% \\
38 vertices & 5\%  & 8 & $-0.000422$ & $[-0.000710,-0.000210]$ & 1.53\% \\
38 vertices & 10\% & 9 & $-0.000208$ & $[-0.000233,-0.000190]$ & 0.75\% \\
\bottomrule
\end{tabular}
\caption{Paired aerodynamic shape-optimization results. Matched denotes matched valid runs; CBD-SA reaches lower final modeled drag in every listed pair.}
\label{tab:crm_perturbed_all}
\end{table}

In the 19--27-vertex study, CBD-SA reduces deformation roughness by 18.2\% at 5\% perturbation and 13.0\% at 10\%, with lower roughness in 7/9 matched pairs at each magnitude. We also compare two FFD baselines, FFD-6 and FFD-30, with 6 and 30 design variables, compared with 192 for CBD and CBD-SA. Under the 38-vertex construction, FFD-30 improves final $C_D$ over FFD-6 by 4.35\% and 4.81\%, while both cage methods reach lower modeled drag than both FFD baselines in every matched valid comparison.

\FloatBarrier
\section{Robot Co-Design}

\FloatBarrier
\subsection{Setup and Hyperparameters}
For our experiments we obtained hyperparameters through a limited random search on the shape design only environments: Affinity $\alpha=0.35$ and Affinity Density $d=0.15$. These hyperparameters were used across experiments. The structural-affinity support is an axis-aligned convex box with 8 vertices and 12 faces, padded by 5\% of the inner-cage bounding-box diagonal.

\FloatBarrier
\subsection{Optimizer Settings}
\begin{table}[!htb]
  \centering
  \centering
  \small
  \setlength{\tabcolsep}{2.5pt}
  \begin{tabular}{lc}
  \toprule
\textbf{Parameter} & \textbf{Value} \\
\midrule
\texttt{actor\_hidden\_dim}  & 128     \\
\texttt{critic\_hidden\_dim} & 256     \\
\texttt{num\_steps}          & 8       \\
\texttt{v\_min}              & $-10.0$ \\
\texttt{v\_max}              & $30.0$  \\
\texttt{policy\_noise}       & $0.0$   \\
\texttt{std\_min}            & $0.05$  \\
\texttt{std\_max}            & $0.5$   \\
\texttt{buffer\_size}        & $500100$ \\
\bottomrule
\end{tabular}
\vspace{5pt}
\caption{FastTD3 parameters.}
\label{tab:fasttd3_settings}
\end{table}
For robot co-design experiments, we used the same optimizer settings for the pretraining stage and for the combined RL+GO stage. These parameters are specified in Table~\ref{tab:fasttd3_settings}. Any settings not specified were taken as default from the FastTD3 repository.

\FloatBarrier
\subsection{Spatula Lift Frames}
\label{app:spatula-lift-comparison}
Here we show snapshots from RL+GO learning to lift a spatula-like object. We observe that the fingertip shapes produced by RL+GO under CBD are very jagged and spiky, leading to poor performance on the environment and an inability to grasp the spatula.
On the other hand, RL+GO under CBD-SA produces a very smooth and wide set of fingertips. Combined with the learned policy, the LEAP hand is able to lift the spatula by sliding underneath and then reorienting it to stay stable.

\newcommand{\spatulaframerow}[2]{%
  \noindent\par\vspace{0.25em}
  \begin{minipage}[t]{0.45\textwidth}
    \centering
    \includegraphics[width=\linewidth]{imgs/slideshow_frames/video_b/frame_#2.jpg}\\
    \small CBD
  \end{minipage}\hfill
  \begin{minipage}[t]{0.45\textwidth}
    \centering
    \includegraphics[width=\linewidth]{imgs/slideshow_frames/video_a/frame_#2.jpg}\\
    \small CBD-SA
  \end{minipage}
  \vspace{1.0em}
}

\begin{figure*}[!htb]
  \centering
  \spatulaframerow{1.00}{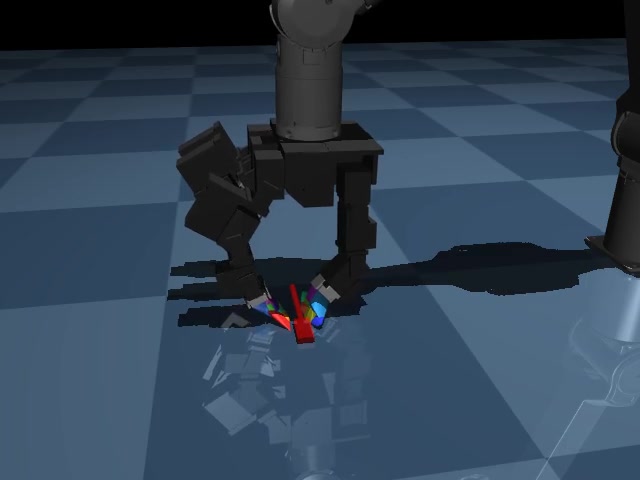}
  \spatulaframerow{2.25}{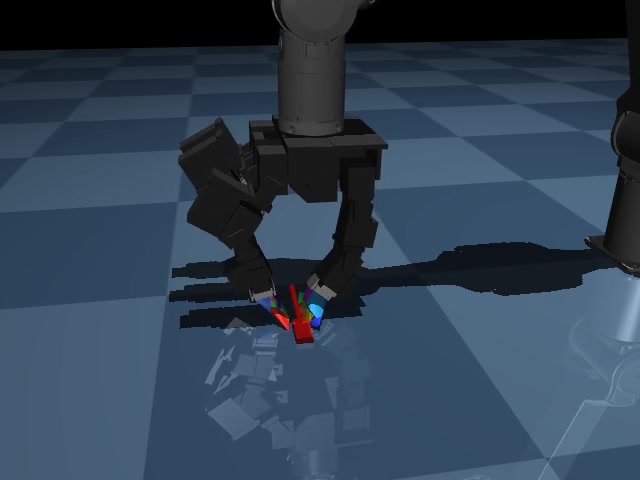}
  \spatulaframerow{3.00}{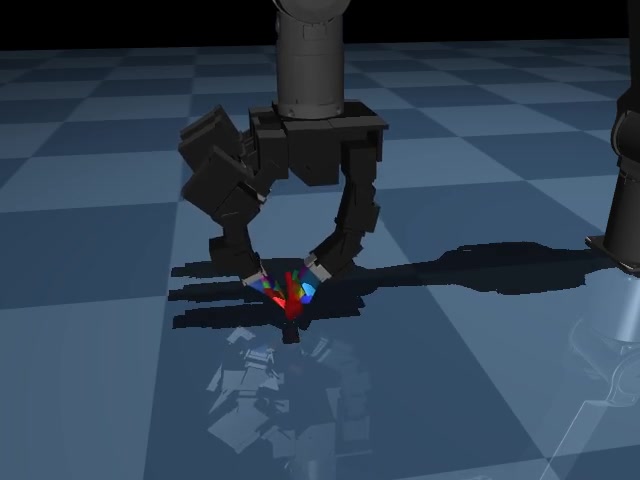}
  \caption{Spatula lift comparison between CBD and CBD-SA with RL+GO optimization.}
  \label{fig:spatula-lift-comparison-1}
\end{figure*}

\FloatBarrier
\section{Compute Resources}
\label{appx:compute}

All Neural Cages and Robot Co-design experiments were carried out on a machine with NVIDIA RTX PRO 6000 Blackwell (96GB VRAM) x 8, 1.5 TB of RAM, and AMD EPYC 9555 64-Core Processor (64-core, 128-thread processor) x2. Each neural cages optimization run took less than 40 minutes. Within the Neural Cages implementation, restoring $\mathcal{L}_{p2f}$ and $\mathcal{L}_{normal}$ increases epoch time by 42.46\% and peak GPU memory by 430 MiB relative to CBD-SA NoReg. Each co-design run took less than 2.5 hours.

All aerodynamic shape optimization experiments were conducted on a machine with 384 vCPU and 768 GB of RAM. Each optimization run took less than 60 minutes.

\FloatBarrier
\section{Licenses}

The ShapeNet/ShapeNetV2 dataset used in the deformation-transfer experiments is used in accordance with the ShapeNet Terms of Use: \url{https://shapenet.org/terms}. These terms permit use of the ShapeNet database for non-commercial research and educational purposes.

The Neural Cages codebase (\url{https://github.com/yifita/deep_cage}) is used under the MIT License.

The aerodynamic shape optimization experiments use OpenAeroStruct (\url{https://github.com/mdolab/OpenAeroStruct}), which is licensed under the Apache License, Version 2.0. OpenAeroStruct builds on OpenMDAO (\url{https://github.com/OpenMDAO/OpenMDAO}), which is also licensed under the Apache License, Version 2.0.

The SLSQP optimizer is used through SciPy, which is distributed under the modified BSD / BSD 3-Clause license. The IPOPT optimizer is licensed under the Eclipse Public License, Version 2.0. Where IPOPT is accessed through pyOptSparse, pyOptSparse is licensed under the GNU Lesser General Public License, Version 3.0.

The Nevergrad optimizer (\url{https://github.com/facebookresearch/nevergrad}) used for gradient-free global optimization is licensed under the MIT License.

The FastTD3 codebase (\url{https://github.com/younggyoseo/FastTD3}) used for reinforcement-learning policy optimization is licensed under the MIT License.

For the LEAP Hand components used in the robotic co-design experiments, LEAP Hand software/code assets are used under the MIT License, the LEAP Hand CAD license, CC BY-NC-SA. The LEAP Hand simulation repository is licensed under the MIT License.

The PiPER robotic arm is used as hardware in the robotic setup. The AgileX piper\_sdk repository is licensed under the MIT License.

The NASA Common Research Model (CRM) geometry/reference case is credited to NASA's CRM data resources. No specific software-style license was identified on the NASA CRM page; we cite the NASA CRM source and use it as a public reference geometry.

\end{document}